\RequirePackage{amsmath,amssymb}
\PassOptionsToPackage{sort}{natbib}
\documentclass[10pt,letterpaper]{sharpa}
\usepackage{lmodern}

\usepackage{amsthm}
\usepackage{xspace,enumitem,tabularx,colortbl}
\usepackage{algorithm,algpseudocode}

\renewcommand{\authorformat}[2][]{%
  \mbox{{\fontsize{11}{14}\selectfont\sffamily\bfseries #2}$^{#1}$}}
\renewcommand{\affiliationformat}[2][]{%
  \mbox{{\fontsize{9}{12}\selectfont\sffamily $^{#1}$#2}}}

\newtheoremstyle{sansserif}{3pt}{3pt}{\normalfont}{}{\sffamily\bfseries}{.}{ }{}
\theoremstyle{sansserif}
\newtheorem{proposition}{Proposition}
\crefname{proposition}{proposition}{propositions}
\Crefname{proposition}{Proposition}{Propositions}
\crefname{algorithm}{algorithm}{algorithms}
\Crefname{algorithm}{Algorithm}{Algorithms}

\makeatletter
\providecommand*{\toclevel@algorithm}{0}
\def\theHALG@line{\theHalgorithm.\arabic{ALG@line}}
\makeatother

\newcommand{\method}{Mem-TTT\xspace}
\newcommand{\pizero}{$\pi_{0.5}$\xspace}
\DeclareMathOperator{\med}{med}
\newcommand{\E}{\mathbb{E}}
\newcommand{\R}{\mathbb{R}}

\definecolor{memoryblue}{HTML}{2F6898}
\definecolor{oursrow}{HTML}{EFF4F9}
\definecolor{tablehead}{HTML}{F1F1F1}
\definecolor{tablecontrol}{HTML}{F7F7F7}
\newcommand{\headrow}{\rowcolor{tablehead}}
\newcommand{\controlrow}{\rowcolor{tablecontrol}}
\newcommand{\oursrow}{\rowcolor{oursrow}}
\newcommand{\methodcell}[1]{\cellcolor{oursrow}\textbf{#1}}
\newcommand{\fullmark}{{\color{memoryblue}\ensuremath{\blacklozenge}}}
\newcommand{\gainmark}[1]{\textbf{#1}}
\newcommand{\best}[1]{\textbf{#1}}
\newcommand{\mmeOurs}{34.7}          
\newcommand{\mmeNoMem}{18.3}         
\newcommand{\mmeIntFreezeVis}{21.5}  
\newcommand{\mmeIntFreezeAct}{19.4}  

\newcommand{\currentRMBNoMem}{23.4}  
\newcommand{\rmbOfficialAvg}{79.5}   
\newcommand{\rmbWindowOne}{78.7}     
\newcommand{\rmbThreeSeedMean}{79.9} 
\newcommand{\rmbThreeSeedSD}{1.3}    
\newcommand{\mmeWindowOne}{34.2}     
\newcommand{\rmbIntFreezeVis}{17.8}  
\newcommand{\rmbIntFreezeAct}{21.7}  

\setcitestyle{numbers,square,comma}

\ifdefined\metadata
  \newcommand{\PaperAuthor}[3]{\appto\authorlist{#1\authorformat[#2]{#3}}}
  \renewcommand{\affiliationlist}{\vspace{-3pt}}
  \newcommand{\PaperAffiliation}[3]{\appto\affiliationlist{#1\affiliationformat[#2]{#3}}}
  \title{Mem-TTT: Plug-in Memory for Pretrained VLAs\\
         via Recall-Supervised Test-Time Training}
\else
  \usepackage{authblk}
  \newcommand{\PaperAuthor}[3]{\author[#2]{#3}}
  \newcommand{\PaperAffiliation}[3]{\affil[#2]{#3}}
  \title{Mem-TTT: Plug-in Memory for Pretrained VLAs via Recall-Supervised Test-Time Training}
  \date{}
\fi

\PaperAuthor{}{1,2}{Mingzhe Ye}
\PaperAuthor{, }{1,2}{Xiao Xia}
\PaperAuthor{, }{1,2}{Congsheng Xu}
\PaperAuthor{, }{1,2}{Zhongxi Chen}
\PaperAuthor{\\[3pt]}{1,3}{Jianbo Zhou}
\PaperAuthor{, }{1,4}{Wenchao Yue}
\PaperAuthor{, }{2}{Xinjun Sheng}
\PaperAuthor{, }{1,2}{Zhijie Deng}

\PaperAffiliation{}{1}{Sharpa Robotics}
\PaperAffiliation{\quad}{2}{Shanghai Jiao Tong University}
\PaperAffiliation{\\[2pt]}{3}{South China University of Technology}
\PaperAffiliation{\quad}{4}{The Chinese University of Hong Kong}

\newcommand{\ProjectWebsiteLink}{\href{https://mem-ttt.github.io/}{\textsf{https://mem-ttt.github.io/}}}
\newcommand{\CorrespondenceLink}{\href{mailto:zhijied@sjtu.edu.cn}{\textsf{zhijied@sjtu.edu.cn}}}
\ifdefined\metadata
  \appto\affiliationlist{\par\vspace{-11pt}}
  \abstract{Test-time training (TTT) equips vision-language-action (VLA) policies with episodic memory by adapting fast weights online.
However, training one flat state end to end with the action loss can overwrite early cues, leaves retention unsupervised, and forces slow sequential optimization.
We present \textbf{\method}, a plug-in, self-supervised multimodal memory bank for pretrained VLAs.
Parallel visual and action streams write frozen-VLM tokens and executed action chunks into bounded short-window fast-weight encoders, then recursively fold completed windows into a hierarchical merge tree in logarithmic depth.
Supervised via timestamped recall to reconstruct past visual features and actions from raw demonstrations, memory learning decouples from policy control, unlocking exact offline causal caching and $7.1\times$ faster policy optimization steps via near-zero-initialized gated adapters.
Across VLA backbones, \method boosts GR00T~N1.7 on RMBench from \currentRMBNoMem\% to \rmbThreeSeedMean\%, nearly doubles \pizero success on RoboMME while outperforming the benchmark's end-to-end TTT baseline by 12.7 points, and lifts four-task mean success on real-robot bimanual dexterous manipulation from 33.5\% to 77.8\%.
}
  \metadata[Project Website]{\ProjectWebsiteLink}
  \correspondence{\CorrespondenceLink}
  \renewcommand{\titlelogo}{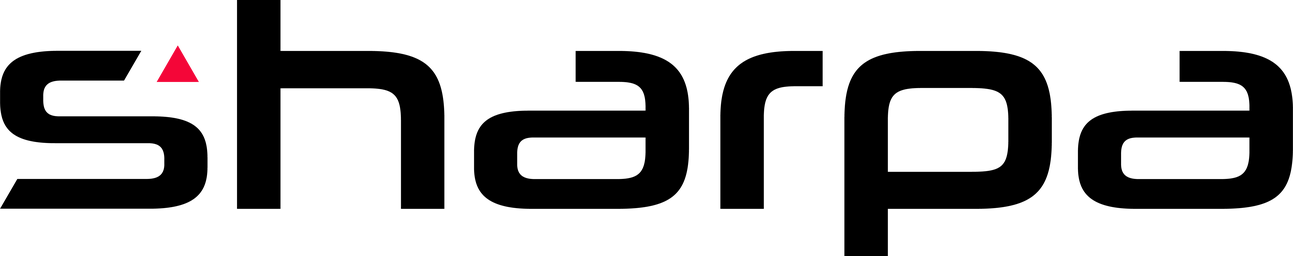}
  \renewcommand{\titlelogowidth}{2.8cm}
  \renewcommand{\titlelogoraise}{-16.7pt}
\fi
\hypersetup{
  pdftitle={Mem-TTT: Plug-in Memory for Pretrained VLAs via Recall-Supervised Test-Time Training},
  pdfauthor={Mingzhe Ye, Xiao Xia, Congsheng Xu, Zhongxi Chen, Jianbo Zhou, Wenchao Yue, Xinjun Sheng, Zhijie Deng}
}

\begin{document}
\maketitle
\ifdefined\metadata
  \vspace{-\baselineskip}
\else
  \begin{abstract}
    Test-time training (TTT) equips vision-language-action (VLA) policies with episodic memory by adapting fast weights online.
However, training one flat state end to end with the action loss can overwrite early cues, leaves retention unsupervised, and forces slow sequential optimization.
We present \textbf{\method}, a plug-in, self-supervised multimodal memory bank for pretrained VLAs.
Parallel visual and action streams write frozen-VLM tokens and executed action chunks into bounded short-window fast-weight encoders, then recursively fold completed windows into a hierarchical merge tree in logarithmic depth.
Supervised via timestamped recall to reconstruct past visual features and actions from raw demonstrations, memory learning decouples from policy control, unlocking exact offline causal caching and $7.1\times$ faster policy optimization steps via near-zero-initialized gated adapters.
Across VLA backbones, \method boosts GR00T~N1.7 on RMBench from \currentRMBNoMem\% to \rmbThreeSeedMean\%, nearly doubles \pizero success on RoboMME while outperforming the benchmark's end-to-end TTT baseline by 12.7 points, and lifts four-task mean success on real-robot bimanual dexterous manipulation from 33.5\% to 77.8\%.

  \end{abstract}
  \noindent\textbf{Project Website:} \ProjectWebsiteLink\par
  \noindent\textbf{Correspondence:} \CorrespondenceLink\par
\fi
\begin{figure}[htbp]
  \centering
  \includegraphics[width=\textwidth]{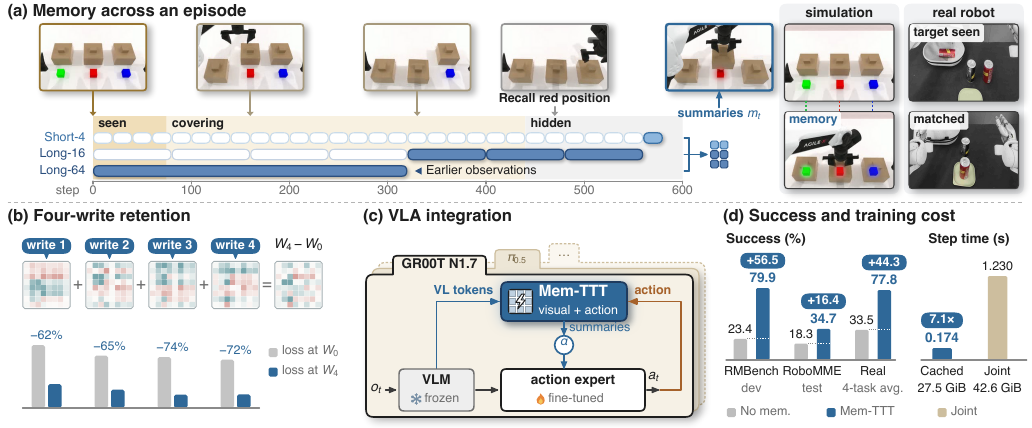}
  \caption{\textbf{\method turns test-time training into a plug-in multimodal memory bank for pretrained VLAs.}
  (a)~Short observation segments are written into fast weights and recursively merged into multi-scale summaries $m_t$ (Short-4, Long-16, Long-64), recovering occluded cues in simulation and on hardware.
  (b)~Four gradient writes accumulate into $W_4 - W_0$, shown via $8\times8$ fast-weight increments at top, while reducing key--value association loss across all four writes at bottom.
  (c)~Adapters attach visual and action memory streams to diverse VLAs such as GR00T~N1.7 and \pizero with the VLM frozen.
  (d)~Task success gains and Stage-3 GR00T policy-step cost: cached causal training cuts step time by $7.1\times$ and peak GPU memory by 35.5\% versus joint memory--policy backpropagation (\cref{sec:cost}).}
  \label{fig:overview}
\end{figure}

\section{Introduction}
\label{sec:intro}

Pretrained vision-language-action (VLA) policies~\citep{rt1,rt2,octo,openvla,pi0,pi05,gr00tn1,xvla} and action-chunking visuomotor models~\citep{dp23,act23,rdt1b} bring strong visual grounding and motor priors to robotic manipulation, yet they condition primarily on the current observation or a short observation window.
In everyday tasks, decisive cues regularly vanish before execution: blocks are covered by cups, a reference object is sealed inside a box, or an action depends on how many times a finger has already pressed a button~\citep{rmbench,robomme}.
Bridging these perceptual gaps requires an episodic memory that retains long-horizon interaction history while preserving the simple fine-tuning workflow of pretrained VLAs.
Existing memory extensions either attend over growing history-token caches whose cost scales with horizon length~\citep{ptp,gmp}, depend on subtask annotations, VLM subgoal planners, or scored top-$k$ keyframes~\citep{mem,memer,rmbench}, or construct specialized spatial and perceptual-cognitive memory banks~\citep{sam2act,memoryvla}.

Test-time training (TTT) stores sequential observations in fast weights updated via online gradient steps~\citep{ttt24}. Robotic formulations train memory through action supervision, using truncated backpropagation in RoboTTT~\citep{robottt} or alternating memory and policy updates in T$^2$Mem~\citep{t2mem}, without an explicit historical-recall objective. This motivates examining write-chain sensitivity, retention, and the cost of sequential memory replay (\cref{sec:theory}).
To overcome this entanglement, we introduce \textbf{\method}, which uses decoupled \emph{recall supervision} on raw demonstrations to build a hierarchically compressed, causally cacheable multimodal memory interface for pretrained VLAs (\cref{fig:overview}).

In \method, bounded four-write Short encoders and position-aligned recursive Long merges summarize recent cues and extended visual and action histories within logarithmic update depth, with fixed spatial anchor queries decoupling storage from policy readout.
Rather than relying on outer action gradients, external teacher models, or memory-specific annotations, slow parameters are trained purely by querying summaries at past timestamps to recall frozen-VLM features and executed action chunks from raw trajectories.
Freezing this recall-trained bank turns online memory into a deterministic causal operator: demonstrations are cached once offline, enabling $7.1\times$ faster shuffled-minibatch policy steps at 64 writes and avoiding joint-BPTT memory overflow at batch 256 through near-zero-initialized gated adapters (\cref{sec:method}).

Plugged into pretrained VLAs across two simulation benchmarks and a bimanual robot with dual 7-DoF arms and high-DoF five-finger dexterous hands, \method lifts GR00T~N1.7 on RMBench from \currentRMBNoMem\% to \rmbThreeSeedMean\%, improves \pizero on RoboMME from \mmeNoMem\% to \mmeOurs\% to surpass the benchmark's end-to-end TTT baseline by 12.7 points, and raises real-robot success by 25--55 percentage points across four memory-dependent tasks.
Our contributions are:
\begin{itemize}[leftmargin=1em,nosep,topsep=2pt]
  \item \textbf{Hierarchically compressed multimodal TTT bank:} Bounded Short fast-weight encoders and position-aligned recursive Long merges that capture short- and long-horizon visual and action histories within logarithmic depth, grounded in our sensitivity diagnostics of fast-weight writers (\cref{sec:theory,sec:bank}).
  \item \textbf{Decoupled recall training and causal caching:} Self-supervised timestamped recall on raw demonstrations\allowbreak---without external teacher models or memory-specific annotations---\allowbreak unlocking exact offline causal caching and $7.1\times$ faster shuffled-minibatch policy optimization steps (\cref{sec:target,sec:training}).
  \item \textbf{Plug-and-play VLA memory interface:} Near-zero-initialized gated adapters conditioning GR00T~N1.7 and \pizero on multi-scale visual and action summaries, validated across simulation, online interventions, and bimanual five-finger dexterous hardware (\cref{sec:experiments}).
\end{itemize}

\section{Related Work}
\label{sec:related}

\paragraph{Generalist VLAs and episodic robot memory.}
Action-chunking~\citep{act23} and diffusion-based visuomotor policies~\citep{dp23,rdt1b} and pretrained VLAs~\citep{rt1,rt2,octo,openvla,pi0,pi05,gr00tn1,xvla,openvlaoft} achieve strong short-horizon control, with modular implementations in frameworks such as StarVLA~\citep{starvla}. RMBench and RoboMME expose sharp failures under visual occlusion, event counting, and delayed reference~\citep{rmbench,robomme}.
To incorporate history, context-extension policies attend over cached past tokens~\citep{ptp,gmp}, condensed multi-frame latents~\citep{condensevla}, or event-selected visual keyframes~\citep{eventvla}. Hierarchical approaches summarize long histories using language memory~\citep{mem}, retrieved keyframes for subtask planning~\citep{memer}, stored subtask outcomes~\citep{rmbench}, or latents trained with semantic QA and keyframe reconstruction~\citep{mem4act}. HiMem-WAM learns high-level skill latents with discovered boundary pseudo-labels~\citep{himemwam}, while other architectures maintain spatial, recurrent, or associative memory states~\citep{smt19,rmt22,sam2act,memoryvla,recat,membodied}.
The plug-in module OnEvoMemory~\citep{onevomemory} adds external memory banks to frozen VLAs through gated adapters and refines them with online rollout outcomes; our hierarchical fast-weight bank learns entirely from offline demonstrations.
Closer to our work, RoboTTT embeds TTT layers inside DiT blocks trained with truncated backpropagation through time~\citep{robottt}; TaskAnchor constructs temporary fast weights from visual history and separately predicts milestone-supervised task states~\citep{taskanchor}; and concurrent T$^2$Mem alternates action-supervised memory learning with policy updates including VLM LoRA adaptation~\citep{t2mem}. \method decouples multimodal TTT writers from policy optimization through self-supervised timestamped recall of frozen-VLM features and executed actions, without auxiliary annotations or action-gradient unrolling.

\paragraph{Fast-weight sequence models and decoupled optimization.}
Fast-weight networks, linear-attention transformers, and selective state-space models compress sequential history into fixed-size associative or recurrent states~\citep{schmidhuber92,ba16,linearattn20,fwp21,rfwp21,mamba,mamba2}.
Building on self-supervised test-time training~\citep{ttt20,ttt24,ttr}, recent sequence architectures introduce delta-rule updates, chunkwise parallelism, and locally optimal inner solvers~\citep{deltanet,gdn,deltaprod,longhorn,titans,lact,mesanet,tnt}.
Basic TTT layers answer current-token queries directly from the evolving fast weights that store history~\citep{ttt24}; TNT combines global/local memories with periodic resets~\citep{tnt}. Our bank probes bounded four-write states into explicit summary tokens for separate policy readout, then recursively merges them with a Long writer shared across scales.
Differentiating outer objectives through long inner-update chains, as in meta-learning~\citep{maml17}, can suffer from recurrent and meta-gradient instability~\citep{pascanu13,mamlpp,metz19,metz21}. Truncated backpropagation reduces memory use but limits long-range credit assignment and introduces gradient bias~\citep{tallec17,shaban19}; persistent evolution strategies correct truncation bias in ES estimates~\citep{pes21}. Replayed recurrent states can become stale as parameters change~\citep{r2d2}.
Inspired by decoupled representation learning in control~\citep{worldmodels,atc,ptp}, freezing recall-trained TTT writers yields exact offline causal caches without unrolled policy backpropagation.

\section{Preliminaries}
\label{sec:theory}

Robotic TTT policies augment pretrained VLAs with fast-weight memory~\citep{robottt,t2mem}. Consider a VLA $\pi_\phi(a_{t:t+H_{\mathrm a}}\mid \mathcal I,o_t,s_t)$---predicting an $H_{\mathrm a}$-step action chunk from language instruction $\mathcal I$, observation $o_t$, and proprioceptive state $s_t$---and a fast-weight network $f_{W_t}$ initialized at slow parameters $W_0$ over an episode of write inputs $x_1,\ldots,x_T$.
At each step $t$, $W_t$ takes an online gradient update of step size $\eta>0$ on a local association loss $\ell_t$ with projections $(\theta_K,\theta_V,\theta_Q)$ before outputting $z_t$ under the outer action loss $\mathcal L_{\mathrm{act}}$:
\begin{equation}
\begin{aligned}
  W_t &= W_{t-1} - \eta\nabla_W\ell_t(W_{t-1}), \\
  \ell_t(W) &= \tfrac{1}{2}\|f_W(\theta_K x_t)-\theta_V x_t\|_2^2, \quad
  z_t = f_{W_t}(\theta_Q x_t).
\end{aligned}
\label{eq:ttt_prelim}
\end{equation}
Although per-step inference cost is constant, motivating diagnostics on manipulation trajectories highlight three structural bottlenecks when coupling one flat chain $W_{0\to T}$ end to end with $\mathcal L_{\mathrm{act}}$ (\cref{fig:stability}), with proofs in the appendix.

\begin{figure}[htbp]
  \centering
  \includegraphics[width=\linewidth]{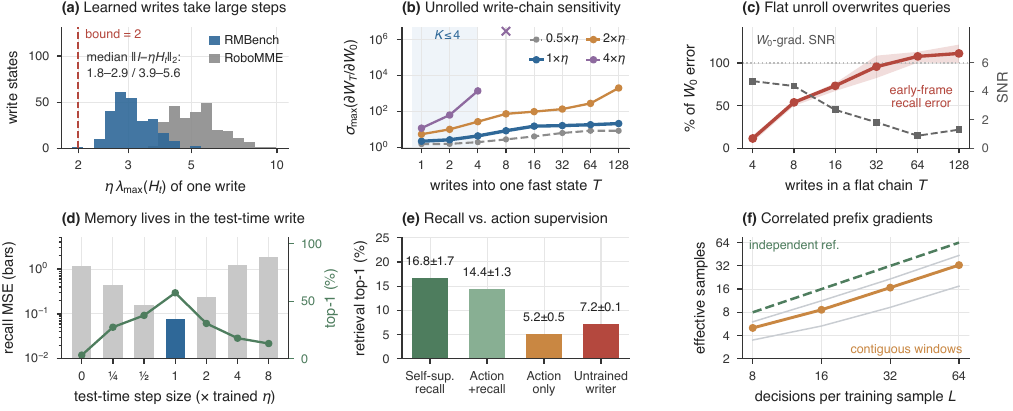}
  \caption{\textbf{Empirical and theoretical diagnostics motivating hierarchical recall-supervised TTT.}
  (a)~Assimilating visual events in a single gradient step on RMBench and RoboMME exhibits learned curvature $\eta\lambda_{\max}(H_t)>2$, exceeding the dashed quadratic contraction bound.
  (b)~Extrapolating a four-write fast state across $T=128$ writes on RMBench compounds $\|\partial W_T/\partial W_0\|_2$ (shading: $T\le 4$; crosses: nonfinite or $>10^6$ at scaled step sizes).
  (c)~Across extended flat chains, early-event recall error at fixed readout queries (solid, 10th--90th percentile band) rises above the unwritten $W_0$ baseline as $W_0$-gradient SNR (dashed) drops.
  (d)~Scaling test-time step size $\eta$ confirms that retained content resides in online gradient writes at $1\times\eta$ (within-batch top-1 retrieval) and vanishes at $0\times\eta$.
  (e)~Under equal-budget training across three seeds on a 512-window candidate pool, direct self-supervised recall yields $3.2\times$ higher top-1 retrieval than action-proxy supervision.
  (f)~Correlation-adjusted effective sample size in contiguous trajectory prefixes lags the independent reference due to temporal gradient correlation.}
  \label{fig:stability}
\end{figure}

\subsection{Write-chain sensitivity and tree depth}
\label{sec:theory_depth}
Differentiating \cref{eq:ttt_prelim} with respect to $W_{t-1}$ for fixed inputs and step size $\eta>0$ gives per-step Jacobian $J_t = I - \eta H_t$, where $I$ is the identity and the symmetric Hessian $H_t = \nabla_W^2\ell_t(W_{t-1})$ has eigenvalues $\{\lambda_i(H_t)\}$ with maximum $\lambda_{\max}(H_t)$~\citep{metz19}:
\begin{proposition}[Write-chain sensitivity]\label{prop:jac}
For differentiable gradient writes with fixed inputs and step size $\eta$,
\begin{equation}
\begin{aligned}
  \frac{\partial W_T}{\partial W_0} &= J_T\cdots J_1, \\
  \left\|\frac{\partial W_T}{\partial W_0}\right\|_2 &\le \prod_{t=1}^{T}\rho_t, \qquad
  \rho_t = \max_i |1 - \eta\lambda_i(H_t)|.
\end{aligned}
\label{eq:prop_jac}
\end{equation}
\end{proposition}
\noindent
Assimilating brief pre-occlusion cues in a single gradient step favors large inner updates rather than slow running averages.
Empirically, fast-weight MLP writers fitted to visual streams on RMBench~\citep{rmbench} and RoboMME~\citep{robomme} operate in a large-step regime where median curvature $\eta\lambda_{\max}(H_t)$ exceeds $2$, yielding step norms $\|J_t\|_2 > 1.8$ (\cref{fig:stability}a).
While subspace cancellations keep $K\le 4$ windows well-conditioned, extrapolating the same four-write state across $T=128$ writes with cycled slot indices raises median $\|\partial W_T/\partial W_0\|_2$ to $21.4$ on RMBench and $159.5$ on RoboMME, cuts $W_0$-gradient SNR by $3.6\times$, and overwrites the fixed readout probes until early-event recall error exceeds the unwritten $W_0$ baseline (\cref{fig:stability}b,c).
Resetting each state after a bounded budget $K$ to emit an explicit summary $r=R(W_K)$ and recursively merging $K$ children across $d=\log_K T$ levels bounds update depth logarithmically:
\begin{proposition}[Depth and event-to-root gain]\label{prop:depth}
In a completed $K$-ary tree spanning $T=K^d$ events with at most $K$ writes per node, any event-to-root path crosses at most $Kd=K\log_K T$ inner updates.
If state-update, write-input, and readout Jacobians have spectral norms bounded by $\rho\ge1$, $b>0$, and $a>0$ at the evaluated inputs, then holding other inputs and slow parameters fixed, the unique path from $x_i$ to root summary $r_{\mathrm{root}}$ satisfies
\begin{equation}
  \left\|\frac{\partial r_{\mathrm{root}}}{\partial x_i}\right\|_2
  \le (ab\rho^{K-1})^d
  = T^{\log_K(ab\rho^{K-1})}.
\label{eq:prop_depth}
\end{equation}
\end{proposition}

\subsection{Outer objectives for historical recall}
\label{sec:theory_recall}
Short tree depth alone does not ensure retention: recursive $K$-ary averaging gives $\|\partial r_{\mathrm{root}}/\partial x_i\|_2 = 1/T$ while erasing temporal order.
Retention instead hinges on the outer objective supervising $(W_0,\theta_{K,V,Q},\eta)$.
Under $\mathcal L_{\mathrm{act}}$, gradients target immediate motor error rather than past scene content and exhibit near-zero cosine similarity of $-0.001$ to $0.006$ with historical feature-recall gradients on Short tokens.
Across four outer objectives trained at equal budget over three seeds on a 512-window candidate pool (\cref{fig:stability}e), an action-prediction proxy reaches only $5.2\%$ top-1 retrieval versus $7.2\%$ for an untrained writer, whereas outer-loop \emph{recall supervision}---querying summaries by timestamp to recover past window features---reaches $16.8\%$.
Scaling test-time step size further verifies that recalled content lives in the online writes: MSE rises from $0.079$ with $\approx58\%$ within-batch retrieval at $1\times\eta$ to $1.173$ at $0\times\eta$ (\cref{fig:stability}d).

\subsection{Prefix correlation and causal caching}
\label{sec:theory_cache}
Backpropagating $\mathcal L_{\mathrm{act}}$ through online writes requires sequential prefix replay, increasing activation memory and step latency. Across 32 consecutive decisions, gradient correlation corresponds to $9.3$--$27.3$ effective samples, compared with $32$ for independent decisions (\cref{fig:stability}f).
Decoupling memory learning from $\mathcal L_{\mathrm{act}}$ removes this coupling: once recall-trained writers are frozen at $\theta^\ast$, online memory becomes a deterministic causal operator $m_{\xi t}=\Phi_{\theta^\ast}(h_{\xi,<t})$ over past trajectory history $h_{\xi,<t}$ under a strict read-before-write boundary.
Precomputing $m_{\xi t}$ once offline therefore yields the exact same memory tensors, losses $\ell_B^{\mathrm{cache}}(\phi;\epsilon)=\ell_B^{\mathrm{replay}}(\phi;\epsilon)$, and policy gradients as online replay for every minibatch $B$, enabling fast globally shuffled VLA fine-tuning without TTT backpropagation.

\section{Method}
\label{sec:method}

Guided by \cref{sec:theory}, \method reformulates test-time training into a decoupled multimodal memory bank $m_t=\Phi_{\theta^\ast}(h_{<t})$ for pretrained VLAs $\pi_\phi(a_{t:t+H_{\mathrm a}}\mid \mathcal I,o_t,s_t,m_t)$, as overviewed in \cref{fig:arch} with full specifications in the appendix.

\begin{figure}[htbp]
  \centering
  \includegraphics[width=\linewidth]{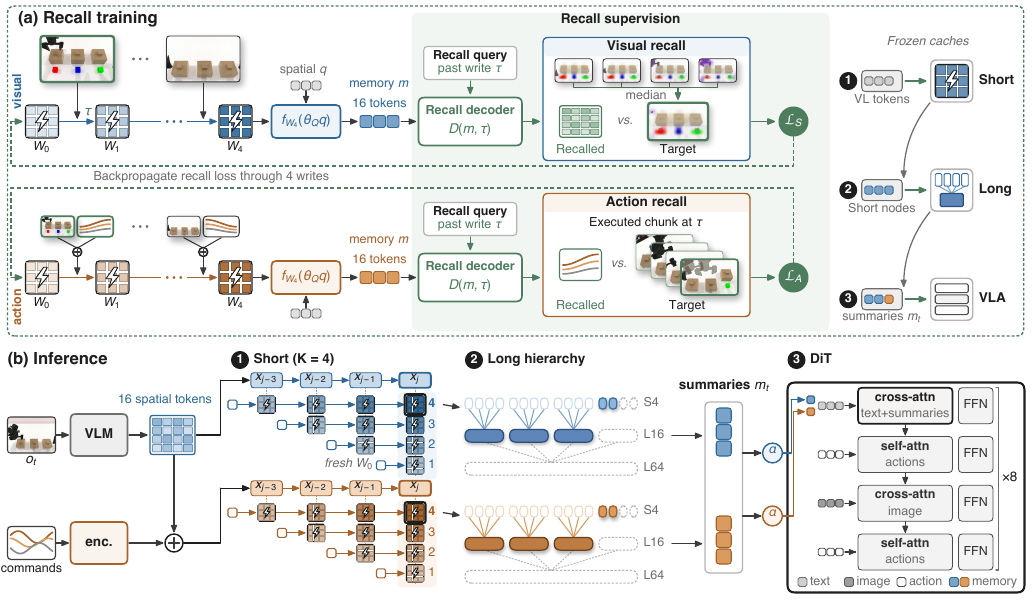}
  \caption{\textbf{Recall-supervised memory training and online hierarchical inference} (illustrated for budget $K=4$ and $M=16$ spatial tokens).
  (a)~$K$ gradient writes produce fast weights $W_K$, probed by spatial anchors $q$ into an $M$-token summary $m$. Training-only decoders query $m$ at past write $\tau$ to recall median frozen-VLM features ($\mathcal L_{\mathrm S}$) or executed command chunks ($\mathcal L_{\mathrm A}$) across three sequential cached stages, and are discarded at deployment.
  (b)~Online, staggered Short states and position-aligned recursive Long merges (S4, L16, L64) supply multi-scale multimodal memory $m_t$ to the VLA through gated adapters.}
  \label{fig:arch}
\end{figure}

\subsection{Recursive TTT memory bank}
\label{sec:bank}
Memory writes occur every $\Delta_{\mathrm w}$ control steps at event index $j$ under a strict \emph{read-before-write} rule so $m_t$ reflects only past events $j < t/\Delta_{\mathrm w}$.
Reusing scene tokens from the VLA's frozen VLM without an external vision encoder, the visual stream average-pools camera patch tokens onto a spatial grid of $M$ cells $z_{j,p}\in\R^{d_{\mathrm{vl}}}$ and projects each cell via $A\in\R^{n\times d_{\mathrm{vl}}}$, LayerNorm, and a learned spatial anchor $q_p\in\R^{n}$:
\begin{equation}
  x_{j,p} = \mathrm{LN}(A z_{j,p}) + q_p \in \R^{n},
  \qquad p \in \{1,\ldots,M\}.
\label{eq:interface}
\end{equation}
In parallel, an action stream projects the normalized command chunk $\bar a_j$ executed since write $j-1$ and modulates it with pooled visual context via a sigmoid-gated MLP to form motor write tokens $x^{\mathrm a}_{j,p}\in\R^{n_{\mathrm a}}$.
The visual Short writer is a two-layer fast-weight MLP $f_W(x)=W_2\,\mathrm{GELU}(W_1x+b_1)+b_2$ of width $n\to 4n\to n$ initialized at $W^{(0)}=W_0$, with an identical $n_{\mathrm a}\to 4n_{\mathrm a}\to n_{\mathrm a}$ writer for actions.
Over $K$ consecutive events $k\in\{0,\ldots,K-1\}$ with slot embeddings $u_k\in\R^{n}$ and $\widetilde x_{j,p}=x_{j,p}+u_k$, $W^{(k)}$ takes $K$ gradient writes across all $M$ cells:
\begin{equation}
\begin{aligned}
  W^{(k+1)} &= W^{(k)} - \eta\nabla_W\ell(W^{(k)};\widetilde x_j), \\
  \ell(W;\widetilde x_j) &= \frac{1}{Mn}\sum_{p=1}^{M}
    \|f_W(\theta_K\widetilde x_{j,p}) - \theta_V\widetilde x_{j,p}\|_2^2.
\end{aligned}
\label{eq:write}
\end{equation}
Sharing $W$ across cells captures cross-region context within the window; after $K$ writes, spatial anchors $q_p$ probe $W^{(K)}$ into an explicit summary $m^{\mathrm S}$ decoupled from policy queries:
\begin{equation}
  m^{\mathrm S}_p = f_{W^{(K)}}(\theta_Q q_p),
  \qquad m^{\mathrm S} \in \R^{M\times n}.
\label{eq:spatial_read}
\end{equation}

Online, $K$ staggered fast states started at $j_0\in\{j-K+1,\dots,j\}$ keep the latest $K$-write summary readable, while non-overlapping blocks starting at $j_0\equiv0\bmod K$ enter the Long hierarchy (\Cref{prop:depth}).
A shared Long merge writer recursively folds $K$ contiguous equal-span child summaries $(c_1,\ldots,c_K)$ into a parent summary $m^{\mathrm L}\in\R^{M\times n}$ across spans $w=K^d$ at tree depths $d\ge 1$ (\cref{fig:arch}b).
Visual merging acts \emph{position-wise} at each cell $p$: child tokens $c_{i,p}$ are projected via a two-layer MLP to inner width $n_{\mathrm L}$, summed with sinusoidal child-index, span, and depth features $\mathrm{Time}(i,w,d)$ plus anchor $q^{\mathrm L}_p$, written via $K$ multi-head gradient steps of \cref{eq:write} into an independent per-cell fast-weight MLP initialized at $W^{\mathrm L}_0$, and linearly projected back to $\R^{n}$:
\begin{equation}
\begin{aligned}
  v_{i,p} &= \mathrm{Project}(c_{i,p}) + \mathrm{Time}(i,w,d) + q^{\mathrm L}_p, \\
  m^{\mathrm L}_p &= \mathrm{Read}_p\!\left(
    \mathrm{Write}(v_{K,p})\circ\cdots\circ
    \mathrm{Write}(v_{1,p})[W^{\mathrm L}_0]\right).
\end{aligned}
\label{eq:merge}
\end{equation}
Independent per-cell states yield a block-diagonal merge Jacobian with $\partial m^{\mathrm L}_p/\partial C_q=0$ for $p\ne q$ across $C_q=(c_{1,q},\dots,c_{K,q})$, preserving spatial alignment while compressing $K^d M$ tokens into $M$ root tokens; the action stream applies the same $K$-ary merge at width $n_{\mathrm a}$.
During execution, a \emph{Lazy} frontier schedule keeps recent summaries at fine temporal resolution and merges the oldest aligned group of $K$ siblings on demand once active Long slots exceed target budget $B_{\mathrm L}$ (\cref{sec:cost}).

\subsection{Self-supervised recall training}
\label{sec:target}

Both streams are trained via self-supervised recall without policy gradients or subtask annotations (\cref{fig:arch}a).
Offline visual targets apply a $(2r+1)$-write centered median $\bar z_{j,p}=\med_{i=j-r}^{j+r} z_{i,p}$, where $r=0$ recovers single-frame VLM features (\cref{sec:exp_target}), followed by training-split standardization with $(\mu,\sigma)$, centering $b_{\mathrm c}$, shared PCA $U\in\R^{d_{\mathrm{vl}}\times n_{\mathrm y}}$, and whitening $s_y$:
\begin{equation}
  y_{j,p} = \mathrm{diag}(s_y)^{-1} U^\top\!
   \left[\mathrm{diag}(\sigma)^{-1}(\bar z_{j,p}-\mu)-b_{\mathrm c}\right]
   \in \R^{n_{\mathrm y}},
\label{eq:target}
\end{equation}
where $j+r$ is accessed only offline when constructing training targets, while online writes use $i\le j$.
In Stage~1, a cross-attention decoder $D_{\mathrm S}(m^{\mathrm S},\tau,p)$ queries visual $m^{\mathrm S}$ at slot $\tau\in\{0,\dots,K-1\}$ and cell $p$ to predict $y_{\tau,p}$ and within-window change $\Delta_\tau y_{\tau,p} = y_{\tau,p} - \frac{1}{K}\sum_{\tau'=0}^{K-1} y_{\tau',p}$ weighted by $\beta>0$:
\begin{equation}
\begin{aligned}
  \mathcal L_{\mathrm S}
  &= \E_{\tau,p}\big\|D_{\mathrm S}(m^{\mathrm S},\tau,p) - y_{\tau,p}\big\|_2^2 \\
  &\quad + \beta\,\E_{\tau,p}\big\|\Delta_\tau D_{\mathrm S}(m^{\mathrm S},\tau,p) - \Delta_\tau y_{\tau,p}\big\|_2^2,
\end{aligned}
\label{eq:recall}
\end{equation}
while an analogous action objective $\mathcal L_{\mathrm A}$ decodes executed chunks $\bar a_\tau$ and $\Delta_\tau \bar a_\tau$ from the action Short summary.
In Stage~2, with Short writers frozen, Long merge writers unroll across episode blocks without detaching children under
\begin{equation}
  \mathcal L_{\mathrm L}
  = \mathcal L_{\mathrm{recon}} + \mathcal L_{\mathrm{align}} + \mathcal L_{\mathrm{hist}},
\label{eq:long_loss}
\end{equation}
where each term averages MSE across cells, channels, and queried timestamps, as detailed in the appendix: $\mathcal L_{\mathrm{recon}}$ sums child-token reconstruction via $D_{\mathrm L}(m^{\mathrm L},i)$ and 8-query recursive recall through $D_{\mathrm L}\to D_{\mathrm S}$; $\mathcal L_{\mathrm{align}}$ aligns the leading $n_{\mathrm y}$ channels of parent, projected-child, and span-64 decoded tokens to span-mean targets; and $\mathcal L_{\mathrm{hist}}$ supervises recall across $1/4\dots4/4$ episode prefix frontiers.
Discarded at deployment, $(D_{\mathrm S}, D_{\mathrm L})$ also render compressed Long nodes directly inspectable.

\subsection{Causal caching and VLA adaptation}
\label{sec:plugin}
\label{sec:training}
In Stage~3, freezing both recall-trained writers at $\theta^\ast$ lets us causally replay every demonstration $\xi$ once under the read-before-write rule to cache $m_{\xi t}$ on disk, enabling flow-matching fine-tuning~\citep{flowmatching23,pi0} on globally shuffled decision minibatches with zero TTT backpropagation.
Each summary token receives continuous age, span, and depth features and enters the frozen-VLM policy via near-zero-initialized gates~\citep{robomme,robottt} (\cref{fig:arch}b).
In GR00T~N1.7~\citep{gr00tn17}, built on the GR00T~N1 architecture~\citep{gr00tn1}, projected visual and action summaries append to cross-attention keys and values in periodic DiT blocks~\citep{dit23} with attention-logit bias $\log|\tanh\alpha|$, whose attention weight $|\tanh\alpha|$ starts near zero at $\alpha_0>0$.
In \pizero, modulator cross-attention~\citep{robomme} in each Gemma action-expert block predicts pre-FFN RMSNorm scale and shift residuals gated by $\tanh g$ from $g_0=0$.
Only the action expert and memory adapters are updated during policy fine-tuning, while fast weights $W$ adapt online at test time.

\section{Experiments}
\label{sec:experiments}

We evaluate \method across two simulation suites, online write interventions, retrained component and target-window ablations, bimanual dexterous hardware, and system profiling, with full per-task tables in the appendix.

\subsection{Benchmarks and experimental setup}
We set write budget and merge arity $K=4$ across Short-4 (S4), Long-16 (L16), and Long-64 (L64), with Long slot budget $B_{\mathrm L}=12$ so each read supplies at most $1+B_{\mathrm L}=13$ active nodes, or up to $208$ visual and $208$ action tokens.
Camera grids of $8\times8$ in \pizero and $8\times10$ in GR00T with $d_{\mathrm{vl}}=2048$ are average-pooled to $M=16$ cells on a $4\times4$ layout, using widths $n=256$, $n_{\mathrm L}=512$, and $n_{\mathrm a}=128$, along with $n_{\mathrm y}=128$ PCA channels over $2r+1=13$ writes for $r=6$ and $\beta=4$.
Memory writes occur every $\Delta_{\mathrm w}=5$ steps in simulation and every $40$-step chunk on hardware, entering every fourth GR00T DiT block across 8 of 32 blocks in simulation and 4 of 16 on hardware from $\alpha_0=10^{-3}$, and \pizero expert blocks from $g_0=0$.
On RMBench~\citep{rmbench}, spanning nine RoboTwin~\citep{robotwin2} tasks with 50 demonstrations each, we fine-tune a shared 50k-step GR00T~N1.7 base per task for 40k steps at batch 256 and evaluate the final checkpoint on 900 fixed episodes plus 1,800 trials across two held-out official scene splits.
On RoboMME~\citep{robomme}, covering 1,600 demonstrations across 16 tasks in four suites, we train a single multi-task \pizero policy with the VLM frozen from the released 80k-step MME-VLA base for 40k steps at batch 256, evaluating 2,400 test and 800 validation rollouts across three 50-episode cohorts per task.
Matched No-memory baselines share identical initialization, data, normalization, and 40k-step budgets without adapters.

\subsection{Main results across VLA backbones}
\label{sec:main}

Under the matched GR00T~N1.7 protocol in \cref{tab:rmbench}, \method lifts success from \currentRMBNoMem\% to \rmbThreeSeedMean$\pm$\rmbThreeSeedSD\% across three independent training runs without the subtask, termination, or QA annotations required by $\dagger$-marked methods, also standing above all published cross-backbone reference scores.
When containers cover objects or identical blocks are swapped mid-episode, the memoryless policy faces ambiguous post-occlusion scenes and Short-only TTT at $30.1\%$ forgets once cues leave its four-write window; by contrast, \method reaches at least $98\%$ on six occlusion, rearrangement, and ranking tasks while leading all matched baselines on \emph{Swap T} and 10-stage \emph{Press Button}, whereas all matched 50-demonstration GR00T variants remain at or below $3\%$ on multi-distractor \emph{Observe \& Pickup}.
Evaluation on RMBench's two held-out official scene splits further reaches \rmbOfficialAvg\% over 1,800 unseen trials.

\begin{table}[htbp]
  \centering
  \small
  \renewcommand{\arraystretch}{1.02}
  \setlength{\tabcolsep}{0.75pt}
  \caption{\textbf{RMBench success rates (\%).}
  Upper block: published cross-backbone results~\citep{rmbench,condensevla,himemwam,mem4act,recat,eventvla} ($\dagger$: auxiliary subtask, termination, or QA supervision; underline: best prior average; QwenOFT and EventVLA averages recomputed from listed task scores in~\citep{eventvla}).
  Lower block: matched GR00T~N1.7 comparison over 100 fixed evaluation trials per task (\fullmark\ \method: mean over three runs, $\pm\rmbThreeSeedSD\%$; bold: strictly best matched score).}
  \label{tab:rmbench}
  \begin{tabularx}{\linewidth}{l*{10}{>{\centering\arraybackslash}X}}
    \toprule
    \headrow Method & Battery & Ranking & Cover & Observe & Press & PutBack & Rearr. & SwapT & SwapB & \textbf{Avg.}$\uparrow$ \\
    \midrule
    \multicolumn{11}{l}{\emph{Published cross-backbone results}} \\
    DP \citep{dp23} & 10 & 10 & 0 & 1 & 0 & 0 & 0 & 20 & 11 & 5.8 \\
    ACT \citep{act23} & 19 & 0 & 0 & 1 & 0 & 0 & 29 & 2 & 2 & 5.9 \\
    \pizero \citep{pi05} & 16 & 6 & 0 & 9 & 0 & 11 & 13 & 15 & 24 & 10.4 \\
    X-VLA \citep{xvla} & 26 & 1 & 2 & 9 & 0 & 18 & 13 & 3 & 16 & 9.8 \\
    QwenOFT \citep{starvla} & 14 & 37 & 0 & 0 & 0 & 0 & 0 & 0 & 0 & 5.7 \\
    MemER$^\dagger$ \citep{memer} & 27 & 0 & 6 & 7 & 0 & 0 & 17 & 7 & 14 & 8.7 \\
    MemoryVLA \citep{memoryvla} & 25 & 12 & 40 & 0 & 0 & 50 & 22 & 9 & 17 & 19.4 \\
    CondenseVLA \citep{condensevla} & 15 & 13 & 22 & 12 & 0 & 46 & 39 & 12 & 35 & 21.6 \\
    HiMem-WAM \citep{himemwam} & 28 & 24 & 19 & 28 & 8 & 32 & 33 & 27 & 38 & 26.3 \\
    Mem-0$^\dagger$ \citep{rmbench} & 28 & 18 & 68 & 4 & 0 & 90 & 89 & 14 & 67 & 42.0 \\
    Mem4Act$^\dagger$ \citep{mem4act} & 29 & 26 & 100 & 22 & 10 & 100 & 99 & 63 & 100 & 61.0 \\
    ReCAT \citep{recat} & 43 & 99 & 50 & 4 & 6 & 100 & 96 & 64 & 100 & 62.4 \\
    EventVLA (VA only) \citep{eventvla} & 35 & 81 & 97 & 21 & 3 & 95 & 96 & 87 & 96 & \underline{67.9} \\
    \midrule
    \multicolumn{11}{l}{\emph{Matched GR00T~N1.7 comparison}} \\
    \controlrow No memory & 68 & 44 & 6 & 3 & 2 & 26 & 46 & 7 & 9 & 23.4 \\
    Short-only TTT ($K{=}4$) & 97 & 49 & 21 & 1 & 3 & 22 & 53 & 11 & 14 & 30.1 \\
    \oursrow \fullmark\ \method & \best{99} & \best{100} & \best{98} & 3 & \best{38} & \best{100} & \best{100} & \best{83} & \best{98} & \best{79.9} \\
    \bottomrule
  \end{tabularx}
\end{table}

On RoboMME~\citep{robomme}, \cref{tab:robomme} compares \method against published perceptual buffers and recurrent/TTT baselines on \pizero.
Perceptual buffers such as FrameSamp-Modul sample $32$ max-pooled $4\times4$ grids totaling $512$ tokens from an episode-wide frame pool and fine-tune the VLM over 80k steps at batch $64$; they excel on \emph{Imitation} at $51.4\%$ versus $19.7\%$, where replaying demonstration videos favors retaining $32$ distinct frames over folding windows recursively.
Unfrozen-VLM recurrent/TTT baselines reach $20.2\text{--}22.4\%$ over 80k steps at batch $16$, whereas causal caching enables 40k-step batch-256 adaptation with the VLM frozen: on the same modulator interface, \method reaches \mmeOurs\%, up from \mmeNoMem\% without memory and $22.0\%$ with TTT-Modul, reaching $28.7\%$ on \emph{Permanence} and surpassing TokenDrop-Modul on \emph{Counting} at $60.2\%$ versus $52.3\%$.

\begin{table}[htbp]
  \centering
  \small
  \renewcommand{\arraystretch}{0.94}
  \setlength{\tabcolsep}{4pt}
  \caption{\textbf{RoboMME test success rates (\%) on \pizero} across the four benchmark suites over 2,400 evaluation rollouts. Upper blocks report published perceptual-buffer and recurrent/TTT baselines \citep{robomme}; the lower block compares matched frozen-VLM policies (bold: best in matched block).}
  \label{tab:robomme}
  \begin{tabularx}{\linewidth}{l*{5}{>{\centering\arraybackslash}X}}
    \toprule
    \headrow Method & Counting & Permanence & Reference & Imitation & \textbf{Avg.}$\uparrow$ \\
    \midrule
    \multicolumn{6}{@{}l}{\emph{Published perceptual-buffer baselines \citep{robomme}}} \\
    TokenDrop (Modul) \citep{robomme} & 52.3 & 26.8 & 34.7 & 38.3 & 38.0 \\
    FrameSamp (Modul) \citep{robomme} & 65.2 & 25.1 & 36.3 & 51.4 & 44.5 \\
    \midrule
    \multicolumn{6}{@{}l}{\emph{Published recurrent and TTT baselines \citep{robomme}}} \\
    RMT (Modul) \citep{rmt22} & 34.1 & 15.8 & 21.1 & 9.7 & 20.2 \\
    TTT (Modul) \citep{ttt24,lact} & 34.5 & 22.2 & 20.4 & 10.7 & 22.0 \\
    TTT (Context) \citep{ttt24,lact} & 36.1 & 21.4 & 22.0 & 9.7 & 22.3 \\
    TTT (Expert) \citep{ttt24,lact} & 36.0 & 23.0 & 19.7 & 10.8 & 22.4 \\
    \midrule
    \multicolumn{6}{@{}l}{\emph{Matched frozen-VLM \pizero comparison}} \\
    \controlrow No memory & 28.0 & 19.3 & 16.5 & 9.3 & 18.3 \\
    Short-only TTT ($K{=}4$) & 30.0 & 23.8 & 23.2 & 12.8 & 22.5 \\
    \oursrow \fullmark\ \method & \best{60.2} & \best{28.7} & \best{30.2} & \best{19.7} & \best{34.7} \\
    \bottomrule
  \end{tabularx}
\end{table}

\FloatBarrier

\begin{figure}[htbp]
  \centering
  \includegraphics[width=\linewidth]{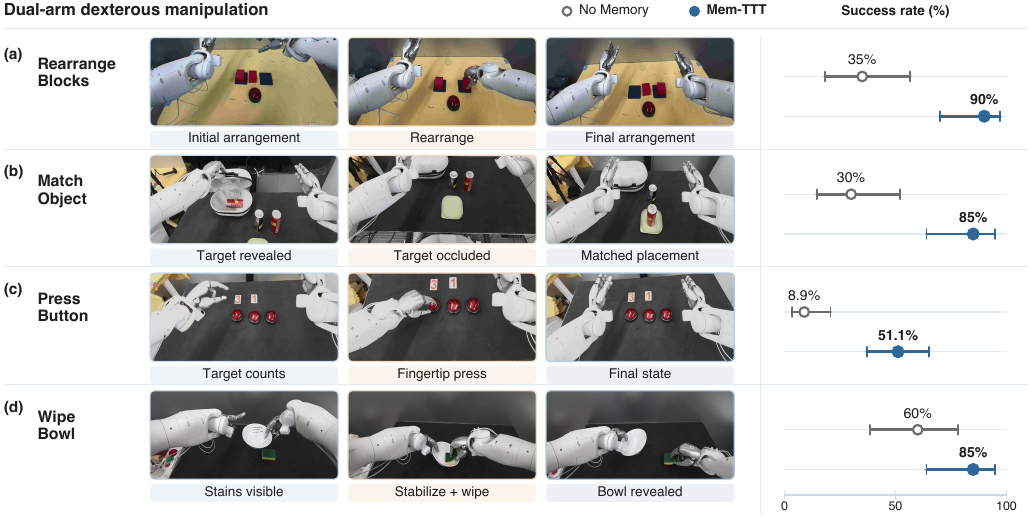}
  \caption{\textbf{Bimanual manipulation on a dual-arm robot with five-finger dexterous hands.}
  Rollout sequences for (a)~spatial block rearrangement, (b)~occluded object-identity matching, (c)~counted finger button presses, and (d)~coordinated bowl stabilization and wiping.
  Right: per-method success rates and two-sided 95\% binomial Wilson score intervals over $n=20, 20, 45, 20$ trials per method for (a)--(d).}
  \label{fig:real}
\end{figure}

\subsection{Memory interventions and ablations}
\label{sec:use}
\label{sec:exp_target}

Intervening on the trained checkpoints at deployment without retraining confirms active reliance on online memory (\cref{tab:interventions}).
Zeroing summaries or disabling online writes with $\eta=0$ feeds constant $W_0$ readouts that shift adapter inputs out of distribution and collapse success, while freezing either visual or action writes causes sharp drops across both backbones.
Critically, substituting in-distribution summaries recorded from an unrelated episode does not cause erratic motion; instead, the policy executes smooth grasps or button sequences toward that episode's remembered target or count, consistent with offline probing where one $16$-token Long-64 root on \emph{Cover Blocks} reconstructs 64-write $4\times4$ grids at $0.050$ MSE versus $0.132$ for an unrelated episode.

\begin{table}[htbp]
  \centering
  \small
  \setlength{\tabcolsep}{6pt}
  \caption{\textbf{Deployment interventions without retraining (\%)} on the default reference \method checkpoints (900 RMBench episodes; 2,400 RoboMME test rollouts).}
  \label{tab:interventions}
  \begin{tabular}{@{}l rr@{}}
    \toprule
    \headrow Backbone & GR00T~N1.7 & \pizero \\
    \headrow Inference condition & RMBench & RoboMME \\
    \midrule
    \oursrow \fullmark\ Normal online \method & \best{79.1} & \best{34.7} \\
    Freeze visual writes & \rmbIntFreezeVis & \mmeIntFreezeVis \\
    Freeze action writes & \rmbIntFreezeAct & \mmeIntFreezeAct \\
    Zeroed summaries & 0.4 & 15.5 \\
    Unrelated episode & 25.6 & 11.9 \\
    All writes off ($\eta=0$) & 0.8 & 13.2 \\
    \bottomrule
  \end{tabular}
\end{table}

Retraining 40k-step policies from scratch clarifies each component's role in \cref{fig:ablations}a, with RoboMME breakdowns in the appendix.
Retaining only the visual branch drives gains on occluded layout and identity tasks such as \emph{Cover Blocks}, \emph{Permanence}, and \emph{Reference}, whereas the action branch, which fuses pooled visual context with executed commands, leads when visual states alias across repetitive motions on \emph{Press Button} and RoboMME \emph{Counting}.
Short-only TTT with $K=4$ improves over No memory on short-horizon cues such as \emph{StopCube} and \emph{VideoRepick}, yet lags full \method once decisive events recede beyond four writes as in \emph{SwingXtimes}.
Across target windows in \cref{fig:ablations}b, single-write direct VLM targets with $r=0$ reach \rmbWindowOne\% on RMBench and \mmeWindowOne\% on RoboMME alongside $79.1\%$ and $34.7\%$ for 13-write targets, showing that nearly the full performance gain persists without temporal median smoothing.

\begin{figure}[htbp]
  \centering
  \includegraphics[width=0.64\linewidth]{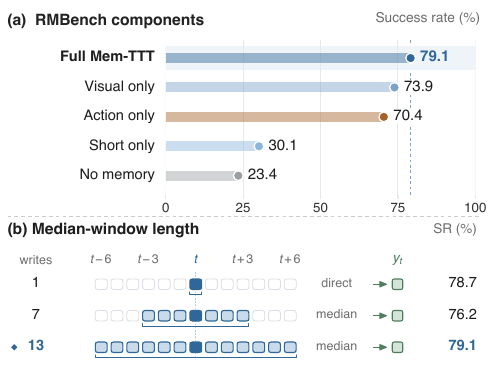}
  \caption{\textbf{RMBench component and target-window ablations} paired with the default reference run (79.1\%, 900 episodes per variant).
  (a)~Retrained policies isolating the visual branch, action branch, and Long summaries.
  (b)~Retrained policies across 1-write direct, 7-write ($\pm3$ median), and 13-write ($\pm6$ median, \fullmark) frozen-VLM targets.}
  \label{fig:ablations}
\end{figure}

\subsection{Dexterous hardware and system efficiency}
\label{sec:real}
\label{sec:cost}
We validate physical manipulation on a bimanual robot with dual 7-DoF arms, two high-DoF five-finger dexterous hands, a head stereo camera, and two wrist fisheye cameras (\cref{fig:real}).
Fast weights reset to $W_0$ at trial start; online, the policy reads the causal frontier before each 40-step relative joint-angle chunk and performs one visual and action write after execution.
Across four partially observable tasks with randomized object and target placements totaling 105 physical rollouts per method, \method lifts unweighted four-task mean success by 44.3 percentage points from 33.5\% to 77.8\% (\cref{fig:real}), recalling initial mat occupancy across symmetric intermediate states in \emph{Rearrange Blocks}, reference-can identity after lid closure in \emph{Match Object}, finger press counts in \emph{Press Button}, and occluded wiping progress in \emph{Wipe Bowl}.

System profiling confirms lightweight training and deployment, with stage-wise GPU hours detailed in the appendix.
At 64 writes and batch 8 on one H20 GPU, cached Stage-3 steps cut GR00T step time from 1.23\,s to 0.174\,s, achieving a $7.1\times$ speedup and 35.5\% lower peak VRAM than joint writer backpropagation while enabling batch-256 training where joint BPTT runs out of memory.
Online, lazy sibling merging holds total dynamic state at 28.5\,MiB, 38.8\% below eager merging, and lowers historical recall error at 7.86\,ms median write latency, whereas a 12-slot sliding FIFO window drops early cues and loses 43--60 points on \emph{Swap Blocks} and \emph{Cover Blocks}.

\FloatBarrier

\section{Conclusion and Future Work}
\label{sec:conclusion}

We presented \method, which reformulates test-time training into a decoupled, self-supervised multimodal memory bank, resolving the update-depth, supervision, and prefix-coupling bottlenecks of end-to-end fast-weight policies.
Combining bounded four-write encoders, position-aligned recursive merges, offline causal caching, and near-zero-initialized gated adapters delivers consistent gains across GR00T~N1.7 on RMBench, \pizero on RoboMME, and bimanual five-finger dexterous hardware.

\paragraph{Limitations and future work.}
Our writers are trained per domain on single-camera $4\times4$ grids without memory dropout, and folding windows recursively rather than buffering individual frames trades dense motion replay such as \emph{Imitation} for bounded online updates.
Extending recall-supervised fast weights to multi-view 3D slots, memory dropout, and tactile streams is a promising direction.

\section*{LLM Usage Statement}
OpenAI ChatGPT, Anthropic Claude, and Google Gemini assisted with research brainstorming, literature exploration, code development and debugging, and manuscript language refinement. The authors reviewed and verified all content.

\FloatBarrier
\setlength{\bibsep}{2pt}
\bibliographystyle{plainnat}
\bibliography{reference}

\clearpage
\appendix
\setcounter{table}{0}
\setcounter{figure}{0}
\setcounter{equation}{0}
\setcounter{algorithm}{0}
\setcounter{proposition}{0}
\renewcommand{\thetable}{S\arabic{table}}
\renewcommand{\thefigure}{S\arabic{figure}}
\renewcommand{\theequation}{S\arabic{equation}}
\renewcommand{\thealgorithm}{S\arabic{algorithm}}
\renewcommand{\theproposition}{S\arabic{proposition}}
\def\theHtable{supp.\arabic{table}}
\def\theHfigure{supp.\arabic{figure}}
\def\theHequation{supp.\arabic{equation}}
\def\theHalgorithm{supp.\arabic{algorithm}}
\def\theHproposition{supp.\arabic{proposition}}
\phantomsection
\addcontentsline{toc}{section}{Appendix}
\section*{Appendix}

This document provides architectural specifications, mathematical proofs, per-task benchmark breakdowns, and real-robot and system profiling details supporting the main paper:
\begin{itemize}[leftmargin=*,nosep,topsep=3pt]
  \item \textbf{\Cref{sec:supp_impl}} details module dimensions, causal execution and three-stage training pseudocode (\Cref{alg:online,alg:training}), Stage~2 loss terms, VLA adapter interfaces, and gate dynamics (\cref{fig:supp_training_gates}).
  \item \textbf{\Cref{sec:supp_theory}} proves the sensitivity, spatial-merge, and temporal-median propositions, visualizes recursive Long-64 spatial recall (\cref{fig:supp_spatial_recall}), and reports inner-loop, supervision, and action-counting diagnostics (\cref{fig:supp_hierarchy}).
  \item \textbf{\Cref{sec:supp_results}} reports complete per-task results on RMBench~\citep{rmbench} and RoboMME~\citep{robomme}, covering multi-run stability, held-out scene splits, retrained component ablations, target-window widths, and online write interventions (\cref{fig:supp_interventions}).
  \item \textbf{\Cref{sec:supp_real_cost}} evaluates matched-visit minibatch ordering, documents the bimanual dexterous hardware setup and task protocols (\cref{fig:supp_real}), and profiles training and online memory overhead (\cref{fig:supp_efficiency}).
\end{itemize}

\section{Architecture, Training, and Gate Dynamics}
\label{sec:supp_impl}

\paragraph{Visual and action inputs and offline recall targets.}
Visual memory reuses scene-camera tokens from the VLA's frozen VLM backbone without introducing an external vision encoder.
In \pizero~\citep{pi05}, the $8\times8$ front-camera grid of 64 tokens is average-pooled in $2\times2$ blocks onto a $4\times4$ grid; in GR00T~N1.7~\citep{gr00tn17}, the 80 head-camera tokens at sequence positions $4\ldots83$, which form an $8\times10$ merged patch grid, are adaptively average-pooled to $4\times4$ by pairing adjacent rows and binning columns into $[0,3)$, $[2,5)$, $[5,8)$, and $[7,10)$.
Each of the $M=16$ spatial cells has $d_{\mathrm{vl}}=2{,}048$ input channels.
Offline visual targets apply a coordinate-wise temporal median over $2r+1=13$ writes with $r=6$ and boundary replication at episode endpoints, standardize channels using training-split statistics $(\mu,\sigma)$, center by $b_{\mathrm c}$, project onto a shared $n_{\mathrm y}=128$-D PCA basis $U$ across cells that retains 93.0\% of variance on RMBench and 96.8\% on RoboMME, and whiten each component by $s_y$ (\cref{tab:supp_modules}).
The action stream combines the executed command chunk $\bar a_j$ with pooled visual context to form motor write tokens $x^{\mathrm a}_{j,p}$, and the Short decoder reconstructs past command chunks $\bar a_\tau$ during pretraining.
All target statistics are fitted on training demonstrations; at deployment, \method consumes only online frozen-VLM tokens and executed commands.

\begin{table}[htbp]
  \centering
  \small
  \caption{\textbf{Multimodal memory bank specifications.} Recall and child decoders ($D_{\mathrm S}, D_{\mathrm L}$) are used only during Stage~1 and Stage~2 pretraining and are discarded before deployment.}
  \label{tab:supp_modules}
  \begin{tabular}{@{}l l@{}}
    \toprule
    \headrow Component & Configuration \\
    \midrule
    Visual input $z_{j,p}$ & $M{\times}d_{\mathrm{vl}} = 16{\times}2048$ \\
    Visual Short $m^{\mathrm S}$ / MLP & $M{\times}n = 16{\times}256$, $256{\to}1024{\to}256$ \\
    Short budget / anchors & $K=4$ writes / $M=16$ anchors $q_p$ \\
    Visual target $y_{j,p}$ & $M{\times}n_{\mathrm y} = 16{\times}128$, 13-write PCA ($r{=}6$) \\
    Visual Long $m^{\mathrm L}$ / width & $16{\times}256$, $n_{\mathrm L}=512$ ($8{\times}64$ heads) \\
    Long merge / anchored & Arity $K=4$, leading $n_{\mathrm y}=128$ channels \\
    Action Short / Long ($n_{\mathrm a}$) & $16{\times}128$, $128{\to}512{\to}128$, Long inner $256$ \\
    \bottomrule
  \end{tabular}
\end{table}

\paragraph{Online execution and three-stage training.}
\Cref{alg:online,alg:training} summarize how we read and write TTT fast weights~\citep{ttt24} and then train the policy on cached representations, following the cached-feature training strategy of prior work~\citep{ptp}.

\begin{algorithm}[htbp]
  \caption{Online Causal Read--Write and Hierarchical Merge}
  \label{alg:online}
  \small
  \begin{algorithmic}[1]
    \Require Frozen VLM, VLA $\pi_\phi$ (chunk $H_{\mathrm a}$; inputs $\mathcal I, o_t, s_t$), frozen writers $(W_0, W^{\mathrm L}_0, \eta, K, \Delta_{\mathrm w}, B_{\mathrm L})$ with anchors $\{q_p\}_{p=1}^M$ and slots $\{u_k\}_{k=0}^{K-1}$.
    \State Initialize active windows $\mathcal B$, stored Long nodes $\mathcal H$, and latest Short $m_\star$ to empty. Set $j\leftarrow0$.
    \For{control step $t = 0, 1, \dots$}
      \If{$t \bmod H_{\mathrm a} = 0$} \Comment{\textbf{Causal Read Before Write}}
        \State Form the read frontier $\mathcal F$ from $\mathcal H$ (budget $B_{\mathrm L}$). Pack $m_t\leftarrow\mathrm{Pack}(m_\star,\mathcal F)$.
        \State Sample action chunk $a_{t:t+H_{\mathrm a}} \sim \pi_\phi(\cdot \mid \mathcal I, o_t, s_t, m_t)$
      \EndIf
      \State Execute $a_t$ in environment
      \If{$(t + 1) \bmod \Delta_{\mathrm w} = 0$} \Comment{\textbf{Scheduled Memory Write}}
        \State Extract $M$-cell VLM grid $z_j$ and executed chunk $\bar a_j$; compute $\{x_{j,p}\}_{p=1}^M$ \Comment{Eq.~\eqref{eq:interface}}
        \State Append fresh state $(j_0{=}j,\, k{=}0,\, W^{(0)}{=}W_0)$ to $\mathcal B$
        \For{each active state $(j_0, k, W^{(k)}) \in \mathcal B$}
          \State Set $\widetilde x_{j,p} \leftarrow x_{j,p} + u_k$; update $W^{(k+1)} \leftarrow W^{(k)} - \eta \nabla_W \ell(W^{(k)}; \widetilde x_j)$ \Comment{Eq.~\eqref{eq:write}}
          \If{$k + 1 = K$} \Comment{Completed $K=4$ writes}
            \State Probe summary $m^{\mathrm S}_p \leftarrow f_{W^{(K)}}(\theta_Q q_p)$ for $p=1,\dots,M$ \Comment{Eq.~\eqref{eq:spatial_read}}
            \State Set $m_\star\leftarrow m^{\mathrm S}$ and remove this window from $\mathcal B$.
            \If{$j_0 \equiv 0 \pmod K$} \Comment{Non-overlapping block}
              \State Push the span-4 block $m^{\mathrm S}$ to $\mathcal H$ and merge each aligned quartet into a span-16 node $m^{\mathrm L}$ \Comment{Eq.~\eqref{eq:merge}}
            \EndIf
          \Else
            \State Store $(j_0,k+1,W^{(k+1)})$ for this window in $\mathcal B$.
          \EndIf
        \EndFor
        \State $j \leftarrow j + 1$
      \EndIf
    \EndFor
  \end{algorithmic}
\end{algorithm}

In Stage~1, probing the updated MLP at the $M=16$ spatial anchors $q_p$ aligns output token $p$ with grid cell $p$, while the training-only Short decoder $D_{\mathrm S}$ cross-attends over all $M$ tokens from a time-and-cell query $(\tau, p)$ with temporal-variation weight $\beta=4$.
In Stage~2, the visual Long writer merges four contiguous equal-span children $\{c_i\}_{i=1}^4$ at span $w\in\{16,64\}$ into $m^{\mathrm L}\in\R^{M\times n}$ without detaching children, and the action branch applies the same $K=4$ architecture at width $n_{\mathrm a}=128$.
All Stage~2 terms in $\mathcal L_{\mathrm L}=\mathcal L_{\mathrm{recon}}+\mathcal L_{\mathrm{align}}+\mathcal L_{\mathrm{hist}}$ (Eq.~\eqref{eq:long_loss} of the main paper) are unscaled mean-squared errors normalized across spatial cells, channels, queried timestamps, and merged nodes or episodes:
(1)~$\mathcal L_{\mathrm{recon}}=\mathcal L_{\mathrm{child}}+\mathcal L_{\mathrm{probe}}$ sums child reconstruction $\|\hat c_{1:4}-c_{1:4}\|_2^2$ averaged over $(4,M,n)$, where $\hat c_i=D_{\mathrm L}(m^{\mathrm L},i)$ and target $c_{1:4}$ is detached, together with 8-query recursive readout via $D_{\mathrm L}\to D_{\mathrm S}$ averaged over $(8,M,n_{\mathrm y})$;
(2)~denoting the parent span mean by $\bar y$ and child-span means by $\bar y^{(i)}$, $\mathcal L_{\mathrm{align}}=\mathcal L_{\mathrm{anc}}+\mathcal L_{\mathrm{in}}+\mathbb{I}[w{=}64]\mathcal L_{\mathrm{dec}}$ sums three coordinate-anchoring terms averaged over merged nodes per episode: $\mathcal L_{\mathrm{anc}}$ anchors $m^{\mathrm L}_{:,1:n_{\mathrm y}}$ to $\bar y$ over $(M,n_{\mathrm y})$ to expose direct span-mean coordinates to the policy, $\mathcal L_{\mathrm{in}}$ aligns live projected child inputs $\mathrm{Project}(c_i)_{:,1:n_{\mathrm y}}$ to $\bar y^{(i)}$ over $(4,M,n_{\mathrm y})$ so inner writes share a consistent coordinate frame across depths, and $\mathcal L_{\mathrm{dec}}$ aligns span-64 decoded children $\hat c_{i,:,1:n_{\mathrm y}}$ to $\bar y^{(i)}$ to prevent intermediate drift during two-hop recursion; and
(3)~$\mathcal L_{\mathrm{hist}}$ averages recursive $(M,n_{\mathrm y})$ recall MSE at four prefix queries $\{0,\lfloor P/4\rfloor,\lfloor P/2\rfloor,\max(0,P{-}5)\}$ evaluated across $1/4,2/4,3/4,$ and $4/4$ episode prefixes of length $P$.
In simulation, each read starts from the stored span-16 nodes and trailing span-4 blocks. If their count exceeds $B_{\mathrm L}=12$, the reader groups the oldest aligned quartets of span-16 nodes into span-64 roots, caching each root for reuse. Benchmark episodes have at most 300 writes, so two merge levels suffice for the read budget.

\begin{algorithm}[htbp]
  \caption{Three-Stage Recall and Cached Policy Training}
  \label{alg:training}
  \small
  \begin{algorithmic}[1]
    \Require Demonstrations $\mathcal D$, frozen VLM, base VLA checkpoint $\phi_0$.
    \State Precompute $4\times4$ VLM grids $\{z_{j,p}\}$ and offline $\pm r$ median PCA targets $\{y_{j,p}\}$ on $\mathcal D$ using training-split statistics (\text{Eq.~\eqref{eq:target}}).
    \State \textbf{Stage 1 (Short recall):} Sample random 4-write windows; unroll $K=4$ FP32 gradient writes from $W_0$; optimize slow parameters $(A, q_{1:M}, u_{0:K-1}, W_0, \theta_{K,V,Q}, \eta)$ and decoder $D_{\mathrm S}$ via $\mathcal L_{\mathrm S}$ (\text{Eq.~\eqref{eq:recall}}).
    \State Freeze Short writers; encode all non-overlapping 4-write blocks across $\mathcal D$.
    \State \textbf{Stage 2 (Hierarchical Long merge):} Over episode block sequences, merge aligned 4-child groups into span-16 and span-64 nodes without detaching children; optimize Long writer and $D_{\mathrm L}$ via $\mathcal L_{\mathrm L}$ (\text{Eq.~\eqref{eq:long_loss}}).
    \State Freeze Long writers; discard training-only decoders $(D_{\mathrm S}, D_{\mathrm L})$.
    \State \textbf{Stage 3 (Causal cache \& VLA adaptation):} Replay every $\xi \in \mathcal D$ causally via \Cref{alg:online} to cache $m_{\xi t}$ on disk.
    \State Initialize gated adapters near zero ($\alpha_0=10^{-3}$ or $g_0=0$); fine-tune $(\phi, \text{adapters})$ for 40k steps on shuffled cached minibatches $B$ under the flow-matching loss $\ell_B^{\mathrm{cache}}$ without TTT backpropagation.
  \end{algorithmic}
\end{algorithm}

\paragraph{Gated VLA adapters and learned gate profiles.}
\Cref{tab:supp_hparams} summarizes adapter and optimization settings across the three platforms.

\begin{table}[htbp]
  \centering
  \footnotesize
  \setlength{\tabcolsep}{3.2pt}
  \renewcommand{\arraystretch}{1.12}
  \caption{\textbf{Training, adapter, and evaluation settings across platforms.} The main Stage~3 policies are fine-tuned for 40,000 steps and evaluated at the final checkpoint.}
  \label{tab:supp_hparams}
  \begin{tabularx}{\linewidth}{@{}>{\raggedright\arraybackslash}p{0.16\linewidth}*{3}{>{\raggedright\arraybackslash}X}@{}}
    \toprule
    \headrow Specification & RMBench (Simulation) & RoboMME (Simulation) & Dexterous Hardware (Real) \\
    \midrule
    VLA backbone & GR00T~N1.7~\citep{gr00tn17} (32-layer DiT) & \pizero~\citep{pi05} (Gemma action expert) & GR00T~N1.7~\citep{gr00tn17} (16-layer DiT) \\
    Base initialization & RoboTwin~\citep{robotwin2} 50k-step base & Released MME-VLA~\citep{robomme} 80k base & Task-specific 30-epoch pretrained base \\
    Demonstrations & 50 per task (450 across 9 tasks) & 1,600 across 16 tasks & 201--451 teleoperated/task (196--437 train, 5--14 val) \\
    Visual write input & Head RGB ($8\times10 \to 4\times4$, 2048-D) & Front RGB ($8\times8 \to 4\times4$, 2048-D) & Left head stereo ($8\times10 \to 4\times4$, 2048-D) \\
    Write stride $\Delta_{\mathrm w}$ & Every 5 control steps & Every 5 control steps & Once per 40-row policy call (20 RGB frames in demonstrations) \\
    Long frontier ($B_{\mathrm L}{=}12$) & On-demand merging (12 slots) & On-demand merging (12 slots) & Hierarchical merging up to depth 2 ($\le 12$ slots) \\
    Gated reader interface & Cross-attn K/V suffix at 8 DiT blocks ($0,4,\dots,28$), $\alpha_0=10^{-3}$ & Pre-FFN RMSNorm scale/shift modulation in expert blocks, $g_0=0$ & Cross-attn K/V suffix at 4 DiT blocks ($0,4,8,12$), $\alpha_0=10^{-3}$ \\
    Stage 1 / Stage 2 budget & Vis.: 40k (batch 256) / 48k (12 ep./GPU); Act.: 24k / 24k & Short: 40k (batch 256); Long: 48k (12 ep./GPU) & Visual: task-specific updates matched by passes over four-write windows (Short) and episodes (Long) \\
    Stage 3 policy budget & 40k steps per task, global batch 256 & 40k steps multi-task, global batch 256 & 40k steps per task, global batch 3,072 \\
    Optimizer \& precision & AdamW (expert $10^{-5}$, adapter $10^{-4}$), cosine, 500-step warmup, BF16 + FP32 master & AdamW cosine schedule, global batch 256, frozen VLM & AdamW, $8\times\text{H20}$ GPUs (384/GPU), clip 1.0 \\
    Evaluation protocol & Fixed test scenes: $9\times100=900$ per run (Runs 1--3); additional test scenes: S0 and S2 (900 each) & Test: $16\times50\times3=2{,}400$ (rollout seeds 7--9) & 4 tasks: 20, 20, 45, 20 physical rollouts per method (105 per method) \\
    \bottomrule
  \end{tabularx}
\end{table}

\begin{figure}[htbp]
  \centering
  \includegraphics[width=\textwidth]{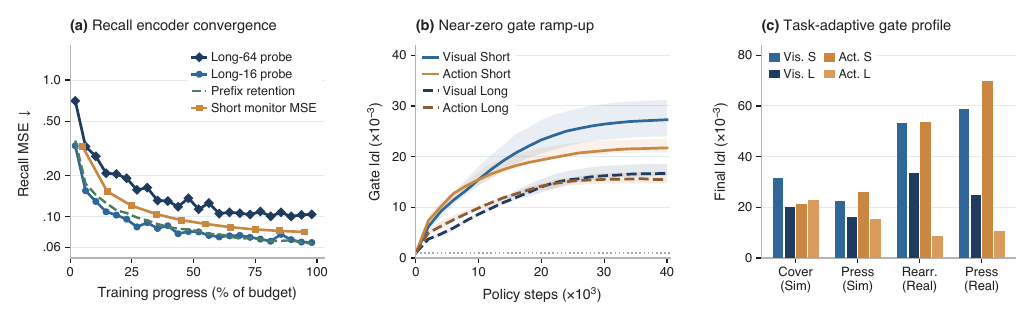}
  \caption{\textbf{Recall convergence and gate trajectories across the three training stages.}
  (a)~Stage~1 visual recall MSE on a fixed subset of 45 training demonstrations reaches $0.077$ in 40k steps (1.59\,h, 5.65\,GiB VRAM), while Stage~2 Long-16 probe, Long-64 probe, and episode prefix-retention losses decrease by more than $30\times$ over 48k steps.
  (b)~During Stage~3 adaptation on RMBench, plotted as the median and interquartile range across nine tasks, all four branch gates start near zero at $|\alpha_0|=10^{-3}$ and open smoothly as the flow-matching loss drops.
  (c)~In simulation, \emph{Cover Blocks} has the largest visual and action Long gates. \emph{Press Button} has the largest action Short gate in both simulation and hardware.}
  \label{fig:supp_training_gates}
\end{figure}

In GR00T~N1.7~\citep{gr00tn17}, the gated suffix adapter normalizes Short and Long tokens separately, adds sinusoidal features of node age, span, and depth together with a learned Short/Long embedding, projects visual tokens from $n$ to $d_{\mathrm{vl}}$ and action tokens from $n_{\mathrm a}$ to $d_{\mathrm{vl}}$, and appends them to cross-attention keys and values in every fourth DiT block with attention-logit bias $\log|\tanh\alpha|$ initialized at $\alpha_0=10^{-3}$, scaling the unnormalized memory attention scores by $|\tanh\alpha_0|\approx 10^{-3}$.
In \pizero~\citep{pi05}, visual ($16\times256$) and action ($16\times128$) summaries are projected to a shared memory width, combined with continuous temporal features and token-index embeddings, and read via memory-as-modulator cross-attention~\citep{robomme} inside each Gemma action-expert block to predict pre-FFN RMSNorm scale and shift residuals gated by $\tanh g$ from $g_0=0$.
As shown in \Cref{fig:supp_training_gates} and \Cref{tab:supp_gates}, starting from $|\alpha_0|=1.0\times10^{-3}$, Stage~3 optimization opens all four gates smoothly alongside a $>100\times$ reduction in flow-matching loss, assigning the largest Long gates to extended occlusion on \emph{Cover Blocks} ($|\alpha_{\mathrm{VL}}|=20.1\times10^{-3}$ and $|\alpha_{\mathrm{AL}}|=22.8\times10^{-3}$) and the largest Short action gates to rapid counting on \emph{Press Button} ($|\alpha_{\mathrm{AS}}|=26.2\times10^{-3}$ in simulation and $70.0\times10^{-3}$ on hardware).

\begin{table}[htbp]
  \centering
  \small
  \caption{\textbf{Learned gate magnitudes ($|\alpha|\times10^3$, initialized at $1.0$)} across RMBench and real-robot policies (bold: column maximum within each platform).}
  \label{tab:supp_gates}
  \begin{tabular}{@{}l rrrr@{}}
    \toprule
    \headrow Task & Vis.\ Short & Vis.\ Long & Act.\ Short & Act.\ Long \\
    \midrule
    \multicolumn{5}{@{}l}{\emph{RMBench simulation (GR00T~N1.7, 8 gated DiT blocks)}} \\
    Cover blocks & \best{31.6} & \best{20.1} & 21.1 & \best{22.8} \\
    Press button & 22.5 & 16.3 & \best{26.2} & 15.5 \\
    Battery try & \best{31.6} & 16.3 & 25.9 & 17.1 \\
    Put back block & 24.0 & 16.7 & 20.9 & 17.1 \\
    Rearrange blocks & 27.7 & 17.2 & 18.8 & 14.5 \\
    Blocks ranking & 23.7 & 18.6 & 18.7 & 16.3 \\
    Swap T & 24.2 & 18.6 & 23.5 & 14.8 \\
    Swap blocks & 27.3 & 15.5 & 21.7 & 14.8 \\
    Observe \& pickup & 31.2 & 14.5 & 22.8 & 14.2 \\
    \midrule
    \multicolumn{5}{@{}l}{\emph{Dexterous real-robot hardware (GR00T~N1.7, 4 gated DiT blocks)}} \\
    Rearrange Blocks & 53.2 & \best{33.5} & 53.5 & 8.5 \\
    Match Object & 23.3 & 9.8 & 17.4 & 0.8 \\
    Press Button & \best{58.9} & 24.7 & \best{70.0} & \best{10.6} \\
    Wipe Bowl & 16.8 & 6.7 & 8.5 & 0.2 \\
    \bottomrule
  \end{tabular}
\end{table}

\FloatBarrier
\section{Proofs and Hierarchy Diagnostics}
\label{sec:supp_theory}

\paragraph{Proofs of main-paper propositions.}
\emph{Proposition~\ref{prop:jac} (Write-chain sensitivity).}
For fixed write inputs and step size $\eta>0$, differentiating $W_t = W_{t-1} - \eta \nabla_W \ell_t(W_{t-1})$ with respect to $W_{t-1}$ gives $J_t = I - \eta H_t$, where $I$ is the identity and the symmetric Hessian $H_t = \nabla_W^2 \ell_t(W_{t-1})$ has real eigenvalues $\{\lambda_i(H_t)\}$~\citep{pascanu13,metz19}.
By the chain rule, $\frac{\partial W_T}{\partial W_0} = J_T \cdots J_1$, and submultiplicativity of the spectral norm yields $\|\partial W_T / \partial W_0\|_2 \le \prod_{t=1}^T \|J_t\|_2 = \prod_{t=1}^T \max_i |1 - \eta \lambda_i(H_t)|$.

\emph{Proposition~\ref{prop:depth} (Depth and event-to-root gain).}
Let a node's $k$-th write be $W^{(k)} = U_k(W^{(k-1)}, c_k)$ for $k\in\{1,\dots,K\}$ initialized at $W^{(0)}=W_0$, and let its anchor readout be $r = R(W^{(K)})$.
Holding the remaining $K-1$ inputs and slow parameters fixed, differentiating $r$ with respect to $c_k$ gives
\begin{equation}
  \frac{\partial r}{\partial c_k} = \frac{\partial R(W^{(K)})}{\partial W^{(K)}}\left(\frac{\partial U_K}{\partial W^{(K-1)}} \cdots \frac{\partial U_{k+1}}{\partial W^{(k)}}\right)\frac{\partial U_k}{\partial c_k},
  \label{eq:supp_event_jac}
\end{equation}
which explicitly incorporates both the write-input Jacobian $\partial U_k/\partial c_k$ and the readout Jacobian $\partial R/\partial W^{(K)}$ beyond state-to-state sensitivity.
When $\|\partial R/\partial W^{(K)}\|_2 \le a$, $\|\partial U_k/\partial c_k\|_2 \le b$, and $\|\partial U_m/\partial W^{(m-1)}\|_2 \le \rho$ with $\rho \ge 1$, submultiplicativity gives $\|\partial r/\partial c_k\|_2 \le a b \rho^{K-k} \le a b \rho^{K-1}$ across at most $K$ inner updates.
Composing \eqref{eq:supp_event_jac} over $d = \log_K T$ tree levels yields at most $K d = K \log_K T$ inner updates and $\|\partial r_{\mathrm{root}}/\partial x_i\|_2 \le (a b \rho^{K-1})^d = T^{\log_K(a b \rho^{K-1})}$ along the unique event-to-root path.

\emph{Complementarity of hierarchical depth and recall supervision.}
Logarithmic update depth $K\log_K T$ limits the length of the update chain. Uniform averaging can still discard temporal order, as $r_{\mathrm{root}}=T^{-1}\sum_{i=1}^T x_i$ illustrates. Timestamped recall supervises reconstruction at specified past writes.
The diagnostics below measure sensitivity and historical reconstruction for the trained writers.

\emph{Causal cache--replay identity.}
With $\theta^\ast$ and preprocessing fixed, matching causal read-before-write boundaries, masks, timestamps, and token packing yield identical memory tensors $m_{\xi t}$ under offline caching and online replay for every minibatch $B$, so $\ell_B^{\mathrm{cache}}(\phi;\epsilon) = \ell_B^{\mathrm{replay}}(\phi;\epsilon)$ and $\nabla_\phi \ell_B^{\mathrm{cache}} = \nabla_\phi \ell_B^{\mathrm{replay}}$ under any minibatch sampling schedule.

\begin{proposition}[Position-wise Long merge]\label{prop:spatial}
With slow parameters and temporal metadata fixed, group the $K$ child summaries at spatial cell $p$ as $C_p=(c_{1,p},\ldots,c_{K,p})$ and write $C=(C_1,\ldots,C_M)$.
In the visual Long merge (Eq.~\eqref{eq:merge} of the main paper), $m^{\mathrm L}_p=F_p(C_p)$ has full Jacobian $J=\partial m^{\mathrm L}/\partial C$ with per-cell blocks $J_p=\partial m^{\mathrm L}_p/\partial C_p$ satisfying
\begin{equation}
 \frac{\partial m^{\mathrm L}_p}{\partial C_{p'}}=0\quad (p\ne p'),\qquad
 \|J\|_2=\max_{p\in\{1,\dots,M\}}\|J_p\|_2,
 \label{eq:supp_spatial_jac}
\end{equation}
and this block-diagonal structure composes across Long levels.
\end{proposition}
\noindent
\emph{Proof.} Each cell $p$ uses an independent fast state with per-cell projections, writes, and anchor reads without cross-cell mixing, giving an $M$-block-diagonal Jacobian inside the merge operator whose composition across tree levels remains $M$-block-diagonal.

\begin{proposition}[Running temporal median]\label{prop:median}
For the coordinate-wise centered temporal median $\bar z_j = \med_{i=j-r}^{j+r} z_i$ over $2r+1$ writes:
(i)~if at most $r$ writes in the window are corrupted by transient arm motion or brief occlusion, each coordinate of $\bar z_j$ remains within the range of the uncorrupted values; and
(ii)~for a persistent scene transition at write $j_0$ away from episode boundaries, a trailing causal median switches at $j_0+r$, lagging by $r$ writes, whereas the centered median switches immediately at $j_0$.
\end{proposition}
\noindent
\emph{Proof.} Because at least $r+1$ of the $2r+1$ samples are uncorrupted, the sample median of each coordinate lies inside their interval, and majority voting across a step change at $j_0$ transitions at $j_0$ for a symmetric window versus $j_0+r$ for a trailing window.

\paragraph{Online frontier memory accounting.}
We measure both peak CUDA memory and dynamic state storage $S_{\mathrm{dyn}}(t)$---the resident byte footprint of all active Short fast-weight states and Long frontier nodes held at step $t$, excluding fixed model parameters.
On-demand merging (\emph{Lazy}) uses 38.8\% less dynamic storage than immediate merging (\emph{Eager}) on the measured episodes.

\paragraph{Inner-loop, hierarchy, and action-counting probes.}

\Cref{tab:supp_inner} summarizes Short-writer curvature and Jacobian norms across 512 four-write windows per benchmark, evaluating curvature via 48-step Lanczos iteration on 64 windows and refining multi-step Jacobian norms via JVP/VJP power iteration on 12 sequences to residual below $0.0104$.
When extrapolating the four-write Short writer to $T=128$ flat writes without state reset across those 12 sequences by cycling slot embeddings $u_{(t-1)\bmod K}$, inner key--value loss on the first four write pairs remains $32.1\text{--}36.0\%$ of the unwritten $W_0$ loss because static background tokens persist; however, anchor-probed recall error across initial slot queries $\tau\in\{0,\dots,3\}$ and targets $y_{0\dots3,p}$ climbs from $11.4\%$ to $111.2\%$ of the $W_0$ baseline on RMBench and from $4.3\%$ to $116.7\%$ on RoboMME as subsequent writes overwrite the $M=16$ spatial probes.
In Fig.~\ref{fig:overview}b of the main paper, we display a representative $8\times8$ submatrix of $W_2\in\R^{n\times 4n}$ normalized to $\pm2.1$\,RMS on \emph{Cover Blocks} alongside the full-state retention losses across all four writes.
Comparing offline-cached and online-replayed memory tensors across four complete demonstrations spanning 587 reads and 4,091 nodes matches all node spans and depths identically, with maximum relative $\ell_2$ difference $2.72\times10^{-6}$ from floating-point reduction order and exact zero cross-cell derivatives $\partial m^{\mathrm L}_p/\partial C_{p'}=0$ for $p\ne p'$.

\begin{table}[htbp]
  \centering
  \small
  \caption{\textbf{Inner-loop Short-writer diagnostics.} Ranges span the four write slots; Jacobian norms are medians over 12 sequences.}
  \label{tab:supp_inner}
  \begin{tabular}{@{}l rr@{}}
    \toprule
    \headrow Quantity & RMBench & RoboMME \\
    \midrule
    Learned step size $\eta$ & 0.220 & 0.321 \\
    Single-write loss ratio & 0.39--0.69 & 0.32--0.62 \\
    Median $\eta\lambda_{\max}$ & 2.70--3.76 & 4.22--6.61 \\
    Maximum $\eta\lambda_{\max}$ & 5.27 & 9.62 \\
    Median $\|I-\eta H_t\|_2$ & 1.81--2.89 & 3.87--5.61 \\
    $\|\partial W_T/\partial W_0\|_2$, $T=4$ & 4.4 & 11.7 \\
    $\|\partial W_T/\partial W_0\|_2$, $T=128$ (extrap.) & 21.4 & 159.5 \\
    $\|\partial W_T/\partial W_0\|_2$, $T=4$ at $2\eta$ & 27.2 & 45.6 \\
    First-4 probe error at $T=4$ (\% $W_0$) & 11.4\% & 4.3\% \\
    First-4 probe error at $T=128$ (\% $W_0$) & 111.2\% & 116.7\% \\
    Recall MSE, learned $\eta$ & 0.079 & 0.088 \\
    Recall MSE, writes off ($\eta=0$) & 1.173 & 0.984 \\
    \bottomrule
  \end{tabular}
\end{table}

\begin{figure}[htbp]
  \centering
  \includegraphics[width=\textwidth]{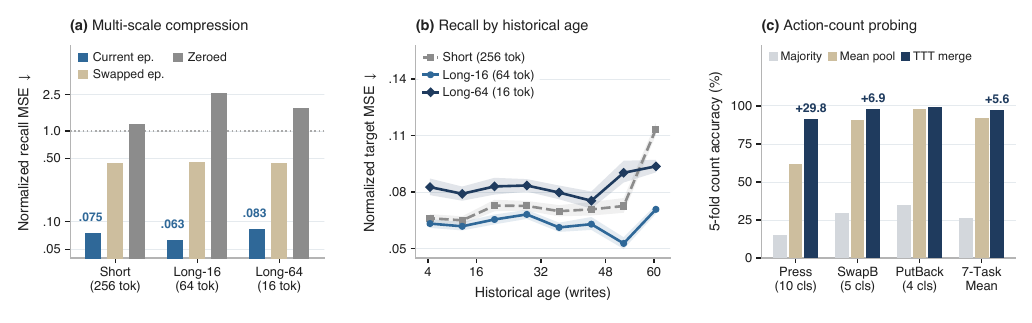}
  \caption{\textbf{Multi-scale compression, recall across age, and action-count probing} on the frozen visual hierarchy.
  (a)~Over matched 64-write spans across 5,248 timestamp queries, a single 16-token Long-64 root compresses 16 Short-4 nodes from 256 tokens by $16\times$ with minimal change in normalized recall MSE ($0.083$ vs.\ $0.075$), whereas swapped-episode memory from an unrelated trajectory and zeroed controls degrade sharply.
  (b)~Across 8-write age bins with 95\% episode-bootstrap intervals, Short-4 degrades on the oldest bin while Long-16 and Long-64 remain flat across all 64 writes.
  (c)~Five-fold cross-episode linear probing of completed grasp and press counts on the 12-slot frontier with on-demand merging shows that TTT merging retains fine-grained event counts substantially better than mean pooling ($91.4\%$ vs.\ $61.6\%$ on 10-class \emph{Press Button}).}
  \label{fig:supp_hierarchy}
\end{figure}

Across 64 episodes totaling 5,248 timestamp queries on the 48k-step Long hierarchy (\cref{fig:supp_hierarchy}a,b), normalized recall MSE is $0.0754$ for 16 Short-4 nodes (256 tokens), $0.0633$ for four Long-16 nodes (64 tokens), and $0.0834$ for a single Long-64 root (16 tokens), compared with $0.4444$ under swapped-episode memory and $1.7898$ when zeroed.

\begin{figure}[htbp]
  \centering
  \includegraphics[width=\textwidth]{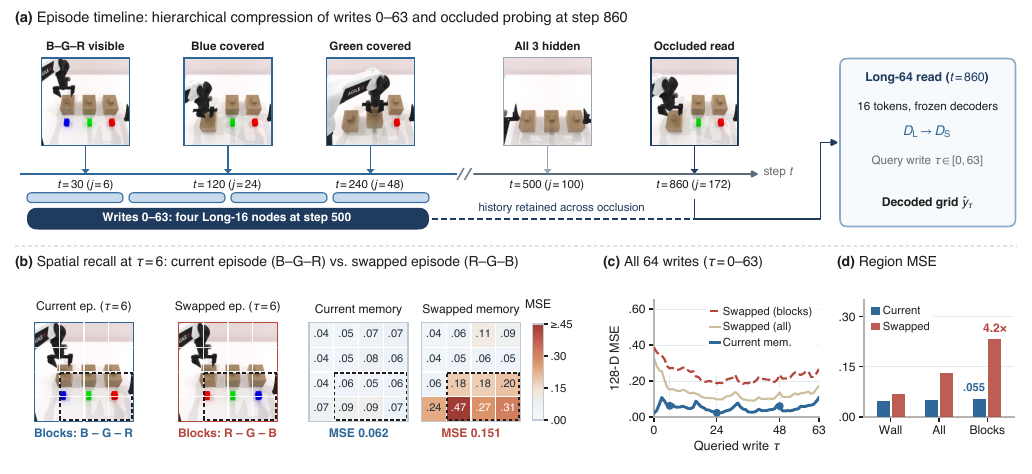}
  \caption{\textbf{Compressing and probing a deployed 16-token Long-64 node under occlusion (\emph{Cover Blocks}).}
  (a)~At full occlusion (step~500), four Long-16 nodes retain writes $0\text{--}63$ (steps $0\text{--}315$). At step~860, frozen decoders $D_{\mathrm L}\to D_{\mathrm S}$ read the 16-token Long-64 root for this history ($16\times$ compression) at query times $\tau\in\{0,\dots,63\}$.
  (b)~At $\tau=6$ (step~30), the 128-D MSE averaged over all $4\times4$ cells is $0.062$ with memory from episode~0 (B--G--R) and $0.151$ with memory from episode~1 (R--G--B). Dashed outlines mark the foreground block cells.
  (c, d)~Across all 64 writes, current memory holds $0.050$ MSE versus $0.132$ overall for swapped memory and achieves $4.2\times$ lower error on block cells ($0.055$ vs.\ $0.233$).}
  \label{fig:supp_spatial_recall}
\end{figure}

\Cref{fig:supp_spatial_recall} visualizes this recursive readout on \emph{Cover Blocks}, comparing episode~0 (B--G--R blocks) against swapped episode~1 (R--G--B blocks) across writes $0\text{--}63$: querying the cached 16-token Long-64 root at decision $j=172$ (step $t=860$) through frozen $D_{\mathrm L}\to D_{\mathrm S}$ reconstructs the $4\times4$ whitened PCA targets at $0.0504$ MSE, which is $2.63\times$ lower than swapped-episode memory overall ($0.1324$), $4.21\times$ lower on the six foreground block cells ($0.0553$ vs.\ $0.2331$, while static wall cells remain $0.0474$ vs.\ $0.0700$), and $37.4\times$ lower than zeroed memory ($1.8828$).
Adding 1--5\% RMS input perturbations yields relative output gains of $1.113$--$1.122$ at Long-64, shifting recall MSE only from $0.0834$ to $0.0850$.

Furthermore, a 5-fold cross-episode logistic probe on PCA-32 pooled features of the 12-slot frontier with on-demand merging across all 450 RMBench demonstrations (\Cref{tab:supp_count_probe} and \cref{fig:supp_hierarchy}c) reaches 97.4\% mean accuracy across the seven multi-count tasks with at least three count classes, versus 91.8\% for span mean pooling, including a 29.8-point margin on 10-class \emph{Press Button} ($91.4\%$ vs.\ $61.6\%$).

\begin{table}[htbp]
  \centering
  \small
  \caption{\textbf{Episodic grasp and button-press count probing (\%)} on the RMBench 12-slot visual frontier with on-demand merging (5-fold cross-episode logistic probe on PCA-32 pooled frontier features).}
  \label{tab:supp_count_probe}
  \begin{tabular}{@{}l r rrrr@{}}
    \toprule
    \headrow Task (event type) & Cls. & Maj. & Time & Pool & \fullmark\ TTT Merge \\
    \midrule
    Press button (presses) & 10 & 15.0 & 73.9 & 61.6 & \best{91.4} \\
    Blocks ranking (grasps) & 17 & 12.3 & 86.2 & 97.2 & \best{97.6} \\
    Cover blocks (grasps) & 7 & 18.4 & 96.0 & 98.4 & \best{98.7} \\
    Swap blocks (grasps) & 5 & 29.3 & 94.3 & 91.0 & \best{97.9} \\
    Battery try (grasps) & 5 & 40.1 & 98.3 & 98.7 & \best{99.0} \\
    Put back block (grasps) & 4 & 34.6 & 97.3 & 98.1 & \best{99.5} \\
    Rearrange blocks (grasps) & 4 & 34.0 & \best{98.6} & 97.7 & 98.0 \\
    Observe \& pickup (grasps) & 2 & 81.5 & \best{98.7} & 94.4 & 95.5 \\
    Swap T (grasps) & 2 & 82.8 & \best{100.0} & 99.9 & 99.4 \\
    \midrule
    \multicolumn{2}{@{}l}{Multi-count mean ($\ge3$ cls.)} & 26.2 & 92.1 & 91.8 & \best{97.4} \\
    \bottomrule
  \end{tabular}
\end{table}

\paragraph{Supervision and policy-gradient diagnostics.}
Fig.~\ref{fig:stability}d of the main paper evaluates the 40k-step Stage~1 writer on 512 four-write windows. Each batch of eight windows supplies 32 target frames for retrieval ($57.2\%$ top-1 at $1\times\eta$, unnormalized MSE $0.079$).
For \cref{tab:supp_supervision} and Fig.~\ref{fig:stability}e, each trained writer runs for 3,000 steps at batch~8. All four conditions, including the untrained control, use fresh decoders trained for 1,500 steps at batch~16 for each of three seeds. Evaluation uses 1,024 sampled test windows in batches of 128, giving 512 target frames per retrieval pool. Recall supervision achieves normalized MSE $0.291$ and top-1 retrieval $16.8\%$, compared with $0.495$ and $5.2\%$ under action-proxy supervision.
In a separate closed-loop comparison (training seed 0), policies with a recall-trained writer, an action-prediction-trained writer, and a frozen random writer succeeded in 712, 608, and 358 of the same 900 RMBench trials, respectively (100 per task).
Recall supervision improved success over action-prediction supervision by 11.56 percentage points (paired 95\% bootstrap CI for these checkpoints, $[9.22,13.89]$).

\begin{table}[htbp]
  \centering
  \small
  \caption{\textbf{Controlled writer supervision comparison} (mean$\pm$SD over 3 seeds; lower is better). Recall MSE is normalized by the training-target mean predictor.}
  \label{tab:supp_supervision}
  \begin{tabular}{@{}l cc@{}}
    \toprule
    \headrow Writer supervision & Norm. recall MSE & Action-proxy MSE \\
    \midrule
    Recall (\method) & $.291\pm.010$ & $.054\pm.006$ \\
    Recall + action proxy & $.301\pm.010$ & $.053\pm.003$ \\
    Action proxy only & $.495\pm.028$ & $.054\pm.004$ \\
    Untrained fast-weight writer & $.400\pm.003$ & $.057\pm.002$ \\
    \bottomrule
  \end{tabular}
\end{table}

\Cref{tab:supp_policy_grads} further reports near-zero cosine similarity between flow-matching policy gradients and historical recall gradients on Short tokens alongside substantial within-episode gradient correlation across $N=32$-step windows. Using Kish's design-effect concept~\citep{kish65}, we compute the correlation-adjusted effective sample size as
\begin{equation}
  \widehat N_{\mathrm{eff}}(N) =
  \frac{N}{1+2\sum_{k=1}^{N-1}(1-k/N)\widehat\rho_k},
  \label{eq:supp_neff}
\end{equation}
where $\widehat\rho_k=\widehat\gamma_k/\widehat\gamma_0$ is the lag-$k$ gradient autocorrelation and $\widehat\gamma_k$ subtracts the squared mean-gradient norm estimated from cross-episode inner products across equally weighted episodes, measuring the variance-equivalent sample size of a contiguous $N$-step gradient average.

\begin{table}[htbp]
  \centering
  \small
  \caption{\textbf{Policy gradient alignment and autocorrelation} on trained GR00T checkpoints (action--recall cosine on Short tokens; lag-1 autocorrelation $\hat\rho_1$; correlation-adjusted effective sample size $\widehat N_{\mathrm{eff}}(32)$).}
  \label{tab:supp_policy_grads}
  \begin{tabular}{@{}l rrr@{}}
    \toprule
    \headrow Task & Cosine & $\hat\rho_1$ & $\widehat N_{\mathrm{eff}}(32)$ \\
    \midrule
    Cover Blocks & $-.001$ & .052 & 21.6 \\
    Press Button & $-.001$ & .106 & 16.7 \\
    Battery & .006 & .251 & 9.3 \\
    Swap T (32 writes) & .002 & .065 & 27.3 \\
    \bottomrule
  \end{tabular}
\end{table}

\FloatBarrier
\section{Per-Task Simulation Results}
\label{sec:supp_results}

\paragraph{RMBench success across training runs and test scenes.}
\Cref{tab:supp_rmb_master} gives per-task success counts for the main RMBench~\citep{rmbench} evaluations in RoboTwin~\citep{robotwin2}, using 100 fixed, unseen test scenes per task (900 episodes per run). We evaluate the final 40k-step checkpoints without selecting checkpoints or hyperparameters from rollout scores.
Across three independent policy training runs initialized with seeds 0--2 (Runs 1--3; Table~\ref{tab:rmbench} of the main paper reports their mean $79.89\pm1.25\%$, while paired ablations and deployment interventions use Run~1), \method reaches $712/900$ (79.11\%), $732/900$ (81.33\%), and $713/900$ (79.22\%).
Whereas RMBench Runs 1--3 measure optimization variance across independent training runs on the fixed 900-episode suite, RoboMME's three cohorts (rollout seeds 7--9 in \cref{tab:supp_robomme_master}) use the official test scenes to measure evaluation rollout variance of a single trained multi-task checkpoint across 2,400 test episodes.
Evaluating the Run~1 policy on RMBench's two held-out official scene splits reaches $724/900$ (80.44\%) on S0 and $707/900$ (78.56\%) on S2, or $79.50\%$ across all $1{,}800$ unseen trials.

\begin{table}[htbp]
  \centering
  \footnotesize
  \setlength{\tabcolsep}{2.5pt}
  \caption{\textbf{RMBench per-task success counts (out of 100)} for matched GR00T~N1.7 policies across component ablations, three independent runs, two additional test sets (S0, S2), and matched-visit ordering schedules (Shuffled vs.\ Contiguous-32).}
  \label{tab:supp_rmb_master}
  \begin{tabular}{@{}l rrrr rrr rr rr@{}}
    \toprule
    & \multicolumn{4}{c}{Component ablations} & \multicolumn{3}{c}{\fullmark\ \method runs} & \multicolumn{2}{c}{\shortstack{Additional\\test scenes}} & \multicolumn{2}{c}{\shortstack{Matched-visit\\order}} \\
    \cmidrule(lr){2-5}\cmidrule(lr){6-8}\cmidrule(lr){9-10}\cmidrule(l){11-12}
    \headrow Task & No mem. & Visual & Action & Short & Run~1 & Run~2 & Run~3 & Off.~S0 & Off.~S2 & Shuf. & Cont32 \\
    \midrule
    Observe and pickup & 3 & 3 & 3 & 1 & 3 & 5 & 2 & 7 & 4 & 6 & 2 \\
    Rearrange blocks & 46 & 100 & 100 & 53 & 100 & 100 & 100 & 100 & 100 & 100 & 100 \\
    Swap blocks & 9 & 90 & 100 & 14 & 100 & 98 & 95 & 100 & 100 & 99 & 99 \\
    Put back block & 26 & 100 & 100 & 22 & 100 & 100 & 100 & 100 & 100 & 100 & 100 \\
    Swap T & 7 & 69 & 74 & 11 & 79 & 87 & 82 & 85 & 81 & 80 & 76 \\
    Press button & 2 & 26 & 34 & 3 & 34 & 45 & 36 & 39 & 30 & 35 & 25 \\
    Battery try & 68 & 97 & 100 & 97 & 100 & 98 & 100 & 99 & 98 & 99 & 100 \\
    Cover blocks & 6 & 95 & 23 & 21 & 96 & 99 & 98 & 95 & 95 & 97 & 91 \\
    Blocks ranking & 44 & 85 & 100 & 49 & 100 & 100 & 100 & 99 & 99 & 99 & 100 \\
    \midrule
    Total / 900 & 211 & 665 & 634 & 271 & 712 & \best{732} & 713 & \best{724} & 707 & \best{715} & 693 \\
    Success rate (\%) & 23.44 & 73.89 & 70.44 & 30.11 & 79.11 & \best{81.33} & 79.22 & \best{80.44} & 78.56 & \best{79.44} & 77.00 \\
    \bottomrule
  \end{tabular}
\end{table}

\begin{figure}[htbp]
  \centering
  \includegraphics[width=\textwidth]{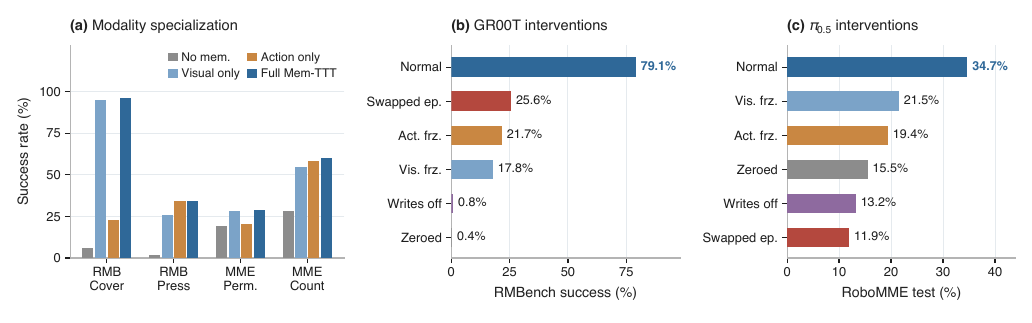}
  \caption{\textbf{Cross-backbone modality specialization and online write interventions.}
  (a)~Across retrained policies on GR00T~N1.7 (RMBench) and \pizero (RoboMME), visual memory leads on occluded spatial tasks (\emph{Cover Blocks}: 95\% vs.\ 23\%; \emph{MME Permanence}: 28.3\% vs.\ 20.2\%), whereas action memory leads on counting (\emph{Press Button}: 34\% vs.\ 26\%; \emph{MME Counting}: 58.2\% vs.\ 55.0\%), and full \method has the highest overall success rate among the retrained component ablations shown.
  (b, c)~Deployment interventions without retraining on RMBench (900 episodes) and RoboMME (2,400 test rollouts) under frozen visual writes, frozen action writes, swapped memory from an unrelated episode, zeroed summaries, and disabled writes ($\eta=0$).}
  \label{fig:supp_interventions}
\end{figure}

\paragraph{Retrained component ablations across backbones.}
Alongside the RMBench ablations in \cref{tab:supp_rmb_master}, we train Short-only, Action-only, and Visual-only \pizero~\citep{pi05} policies on RoboMME~\citep{robomme} for 40k steps from the same 80k-step checkpoint at global batch 256 (\cref{fig:supp_interventions}a and \cref{tab:supp_mme_suites,tab:supp_robomme_master}).

\begin{table}[htbp]
  \centering
  \small
  \caption{\textbf{RoboMME retrained component and target-window summary (\%)} across suites (600 test rollouts per suite, 2,400 total).}
  \label{tab:supp_mme_suites}
  \begin{tabular}{@{}l rrrr r@{}}
    \toprule
    \headrow Configuration & Count & Perm. & Ref. & Imit. & \textbf{Test}$\uparrow$ \\
    \midrule
    \controlrow No memory & 28.0 & 19.3 & 16.5 & 9.3 & 18.3 \\
    Short only & 30.0 & 23.8 & 23.2 & 12.8 & 22.5 \\
    Action only & 58.2 & 20.2 & 21.3 & 10.0 & 27.4 \\
    Visual only & 55.0 & 28.3 & 28.2 & \best{22.3} & 33.5 \\
    \method (1-write) & 59.7 & 28.0 & 28.7 & 20.5 & 34.2 \\
    \oursrow \fullmark\ \method (13-write) & \best{60.2} & \best{28.7} & \best{30.2} & 19.7 & \best{34.7} \\
    \bottomrule
  
  \end{tabular}
\end{table}

Both backbones exhibit a consistent modality split: Action-only leads on \emph{Counting}, including \emph{SwingXtimes}, whereas Visual-only leads on \emph{Permanence}, \emph{Reference}, and \emph{Imitation}.
Removing the Long merge hierarchy produces the same ordering across both benchmarks: Short-only improves over No memory across all four RoboMME suites at $539/2{,}400$ ($22.46\%$) and helps on recent-horizon tasks such as \emph{StopCube}, \emph{VideoRepick}, \emph{PickHighlight}, and \emph{ButtonUnmask}, yet trails full \method at $832/2{,}400$ ($34.67\%$) overall, with lower success on \emph{SwingXtimes} and \emph{PickXtimes}.
When both \emph{Cover Blocks} and \emph{Press Button} were excluded from visual PCA fitting and Short/Long visual writer pretraining, policies trained on their own task demonstrations succeeded in 96/100 and 35/100 trials, respectively, compared with 96/100 and 34/100 for the matched policies using the default visual writer.

\paragraph{Target-window length and online interventions.}
On RMBench (\cref{tab:supp_target_rmb}), 1-, 7-, and 13-write frozen-VLM targets reach $708/900$ (78.67\%), $686/900$ (76.22\%), and $712/900$ (79.11\%) under matched global batch 256, with paired 95\% confidence intervals of $[-1.22,2.11]$ percentage points between 13-write and 1-write targets and $[1.11,4.67]$ between 13-write and 7-write targets.

\begin{table}[htbp]
  \centering
  \small
  \caption{\textbf{RMBench per-task visual target-window study} across 1, 7, and 13 frozen-VLM writes (successes out of 100; bold: strictly best).}
  \label{tab:supp_target_rmb}
  \begin{tabular}{@{}l rrr@{}}
    \toprule
    \headrow Task & 1 write & 7 writes & \fullmark\ 13 writes \\
    \midrule
    Observe and pickup & \best{4} & 3 & 3 \\
    Rearrange blocks & 100 & 100 & 100 \\
    Swap blocks & 100 & 99 & 100 \\
    Put back block & 100 & 100 & 100 \\
    Swap T & 79 & 79 & 79 \\
    Press button & 28 & 33 & \best{34} \\
    Battery try & 100 & 100 & 100 \\
    Cover blocks & \best{98} & 74 & 96 \\
    Blocks ranking & 99 & 98 & \best{100} \\
    \midrule
    Total / 900 & 708 & 686 & \best{712} \\
    Success rate (\%) & 78.67 & 76.22 & \best{79.11} \\
    \bottomrule
  \end{tabular}
\end{table}

On RoboMME (\cref{tab:supp_mme_suites,tab:supp_robomme_master}), 1-write targets reach $821/2{,}400$ ($34.21\%$ test) alongside $832/2{,}400$ ($34.67\%$ test) for 13-write targets.

\begin{table}[htbp]
  \centering
  \footnotesize
  \setlength{\tabcolsep}{2.5pt}
  \caption{\textbf{RoboMME 16-task test results (successes out of 150 per task across rollout seeds 7--9).} Horizontal blocks group Counting, Permanence, Reference, and Imitation. Columns 2--7 report separately trained \pizero policies; columns 8--12 apply test-time interventions to the 13-write \method checkpoint.}
  \label{tab:supp_robomme_master}
  \begin{tabular}{@{}l r rrr rr rrrrr@{}}
    \toprule
    & Base & \multicolumn{3}{c}{Retrained components} & \multicolumn{2}{c}{\method window} & \multicolumn{5}{c}{\shortstack{Deployment interventions\\on 13-write \method}} \\
    \cmidrule(lr){2-2}\cmidrule(lr){3-5}\cmidrule(lr){6-7}\cmidrule(l){8-12}
    \headrow Task & No mem. & Short & \shortstack{Act.\\only} & \shortstack{Vis.\\only} & 1-write & \fullmark\ 13-write & \shortstack{Frz.\\Vis.} & \shortstack{Frz.\\Act.} & Zeroed & Swapped & \shortstack{Writes\\off} \\
    \midrule
    BinFill & 50 & 51 & \best{63} & 62 & 61 & 58 & 51 & 29 & 32 & 23 & 33 \\
    PickXtimes & 54 & 52 & 109 & 103 & 108 & \best{110} & 79 & 50 & 49 & 51 & 42 \\
    SwingXtimes & 58 & 55 & \best{140} & 123 & 136 & 136 & 92 & 43 & 52 & 24 & 20 \\
    StopCube & 6 & 22 & 37 & 42 & 53 & \best{57} & 7 & 22 & 5 & 19 & 11 \\
    \midrule
    VideoUnmask & 43 & 40 & 40 & \best{64} & \best{64} & 63 & 31 & 58 & 35 & 35 & 27 \\
    ButtonUnmask & 34 & \best{43} & 32 & 42 & 41 & 39 & 31 & 30 & 35 & 21 & 24 \\
    VideoUnmaskSwap & 28 & 39 & 36 & 34 & 35 & \best{40} & 23 & 33 & 21 & 25 & 18 \\
    ButtonUnmaskSwap & 11 & 21 & 13 & \best{30} & 28 & \best{30} & 2 & 11 & 1 & 6 & 6 \\
    \midrule
    PickHighlight & 14 & 32 & 28 & 38 & 43 & \best{47} & 26 & 12 & 14 & 2 & 13 \\
    VideoRepick & 1 & 20 & 34 & \best{39} & 31 & \best{39} & 27 & 3 & 0 & 12 & 1 \\
    VideoPlaceButton & 47 & 46 & 37 & 51 & \best{53} & 46 & 40 & 41 & 31 & 20 & 31 \\
    VideoPlaceOrder & 37 & 41 & 29 & 41 & 45 & \best{49} & 38 & 33 & 37 & 20 & 38 \\
    \midrule
    MoveCube & 34 & 47 & 38 & \best{76} & 66 & 74 & 48 & 69 & 37 & 9 & 34 \\
    InsertPeg & 2 & 3 & 0 & 5 & \best{7} & 5 & 2 & 1 & 1 & 0 & 0 \\
    PatternLock & 15 & 18 & 10 & \best{24} & 21 & 16 & 7 & 15 & 16 & 9 & 5 \\
    RouteStick & 5 & 9 & 12 & \best{29} & \best{29} & 23 & 13 & 16 & 5 & 10 & 13 \\
    \midrule
    Test total / 2,400 & 439 & 539 & 658 & 803 & 821 & \best{832} & 517 & 466 & 371 & 286 & 316 \\
    Test rate (\%) & 18.29 & 22.46 & 27.42 & 33.46 & 34.21 & \best{34.67} & 21.54 & 19.42 & 15.46 & 11.92 & 13.17 \\
    \bottomrule
  \end{tabular}
\end{table}

\Cref{tab:supp_robomme_master,tab:supp_rmb_interventions} and \cref{fig:supp_interventions}b,c report per-task deployment intervention counts across all 16 RoboMME and 9 RMBench tasks.
With memory from another episode sharing the same goal, the policy achieved the original goal in 92/100 \emph{Cover Blocks} trials and 100/100 \emph{Put Back Block} trials.
With memory from a different-goal episode, original-goal success fell to 0/100 on each task: the robot followed the donor's entire covering order in 80/100 \emph{Cover Blocks} trials and returned the block to the donor's location in 100/100 \emph{Put Back Block} trials.

\begin{table}[htbp]
  \centering
  \small
  \caption{\textbf{RMBench per-task deployment interventions} on the paired Run~1 \method checkpoint (successes out of 100; bold: strictly best).}
  \label{tab:supp_rmb_interventions}
  \begin{tabular}{@{}l r rrrrr@{}}
    \toprule
    \headrow Task & \fullmark\ Norm. & Frz.V & Frz.A & Zero & Swap. & Off \\
    \midrule
    Observe & 3 & 3 & 3 & 0 & 3 & 3 \\
    Rearrange & \best{100} & 0 & 33 & 0 & 28 & 0 \\
    SwapB & \best{100} & 0 & 85 & 0 & 33 & 0 \\
    PutBack & \best{100} & 2 & 5 & 0 & 30 & 0 \\
    SwapT & \best{79} & 42 & 22 & 4 & 4 & 2 \\
    Press & \best{34} & 0 & 8 & 0 & 9 & 0 \\
    Battery & \best{100} & 95 & 32 & 0 & 66 & 2 \\
    Cover & \best{96} & 18 & 6 & 0 & 11 & 0 \\
    Ranking & \best{100} & 0 & 1 & 0 & 46 & 0 \\
    \midrule
    Total / 900 & \best{712} & 160 & 195 & 4 & 230 & 7 \\
    Rate (\%) & \best{79.1} & 17.8 & 21.7 & 0.4 & 25.6 & 0.8 \\
    \bottomrule
  \end{tabular}
\end{table}

\FloatBarrier
\paragraph{Recent visual and action history on RoboMME.}
The baseline keeps the 13 most recent visual events and the 13 most recent executed-action events, or all events available in a stream if fewer than 13 have occurred.
With the VLM frozen for 40k policy updates at global batch 256, Mem-TTT and the recent-history baseline succeed in 832/2,400 and 613/2,400 test rollouts, respectively (\cref{tab:supp_robomme_fifo}).

\begin{table}[htbp]
  \centering\small
  \caption{\textbf{RoboMME success with recent history alone or Mem-TTT.} The recent-history baseline retains up to 13 events per stream. Entries are successful rollouts. Each task has 150 test rollouts (50 per seed 7, 8, and 9). Suite rows sum the four tasks below them.}
  \label{tab:supp_robomme_fifo}
  \begin{tabular}{@{}lrr@{\hspace{1.5em}}lrr@{}}
    \toprule
    Task or suite & Recent 13 & Mem-TTT & Task or suite & Recent 13 & Mem-TTT \\
    \midrule
    \textit{Counting} & 212 & 361 & \textit{Permanence} & 141 & 172 \\
    \enspace BinFill & 47 & 58 & \enspace VideoUnmask & 58 & 63 \\
    \enspace PickXtimes & 63 & 110 & \enspace ButtonUnmask & 40 & 39 \\
    \enspace SwingXtimes & 91 & 136 & \enspace VideoUnmaskSwap & 27 & 40 \\
    \enspace StopCube & 11 & 57 & \enspace ButtonUnmaskSwap & 16 & 30 \\
    \midrule
    \textit{Reference} & 135 & 181 & \textit{Imitation} & 125 & 118 \\
    \enspace PickHighlight & 31 & 47 & \enspace MoveCube & 73 & 74 \\
    \enspace VideoRepick & 23 & 39 & \enspace InsertPeg & 2 & 5 \\
    \enspace VideoPlaceButton & 38 & 46 & \enspace PatternLock & 36 & 16 \\
    \enspace VideoPlaceOrder & 43 & 49 & \enspace RouteStick & 14 & 23 \\
    \midrule
    Total & 613 & 832 & Success (\%) & 25.54 & 34.67 \\
    \bottomrule
  \end{tabular}
\end{table}

\paragraph{Decoupled and end-to-end policy training.}
We train each policy for up to 20k updates (global batch 64, seed 0). For each task, completed runs share the same 300 unseen test scenes: 100 fixed scenes and 100 each from S0 and S2, with no overlap.
On Cover Blocks, cached training succeeds in 247/300 trials (82.3\%) versus 53/300 (17.7\%) for end-to-end (E2E) training with action and auxiliary recall losses. Training with action loss alone stopped before 20k updates after a non-finite loss. The 98\% reported in the main paper averages three runs on the 100 fixed scenes at 40k updates and batch 256.
On Press Button, cached training succeeds in 50/300 trials (16.7\%), versus 20/300 (6.7\%) for action loss alone and 31/300 (10.3\%) with the auxiliary recall loss.

\section{Real-Robot Protocol and System Efficiency}
\label{sec:supp_real_cost}

\begin{figure}[htbp]
  \centering
  \includegraphics[width=\textwidth]{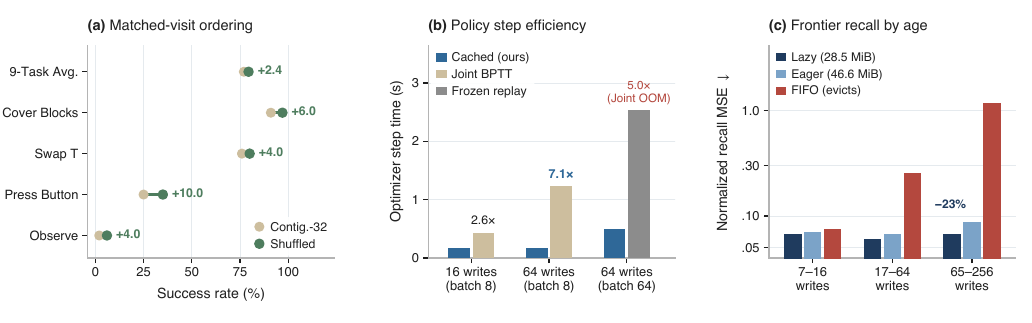}
  \caption{\textbf{Matched-visit ordering, Stage~3 step latency, and online frontier recall by age.}
  (a)~Over the same multiset of $10{,}240{,}000$ cached decision visits per task, shuffling decisions across episodes (\emph{Shuffled}, 79.44\%) outperforms chronological 32-step blocks (\emph{Contiguous-32}, 77.00\%), led by \emph{Press Button} ($+10.0$) and \emph{Cover Blocks} ($+6.0$).
  (b)~Single-H20 GR00T~N1.7 Stage~3 step time: cached training is $7.1\times$ faster at 64 writes than joint writer--policy BPTT and $5.0\times$ faster than detached prefix replay, where joint BPTT runs out of memory.
  (c)~Across 450 RMBench streams, On-demand merging uses 38.8\% less dynamic memory than immediate merging ($28.5$ vs.\ $46.6$\,MiB) and lowers recall MSE by 23\% on 65--256-write history, whereas the 12-slot sliding buffer (FIFO in the plot) discards the oldest spans.}
  \label{fig:supp_efficiency}
\end{figure}

\paragraph{Batch ordering with matched example visits.}
To separate minibatch ordering from example selection, we compare two schedules over the identical multiset of $10{,}240{,}000$ cached decision visits per task across 40,000 steps at batch 256 from shared initialization on Run~1: one schedule shuffles decisions across episodes and the other groups 32 consecutive steps from the same episode (\cref{fig:supp_efficiency}a and \cref{tab:supp_rmb_master}), lifting 900-episode success from $77.00\%$ ($693/900$) to $79.44\%$ ($715/900$, $+2.44$ percentage points).

\begin{figure}[htbp]
  \centering
  \includegraphics[width=\textwidth]{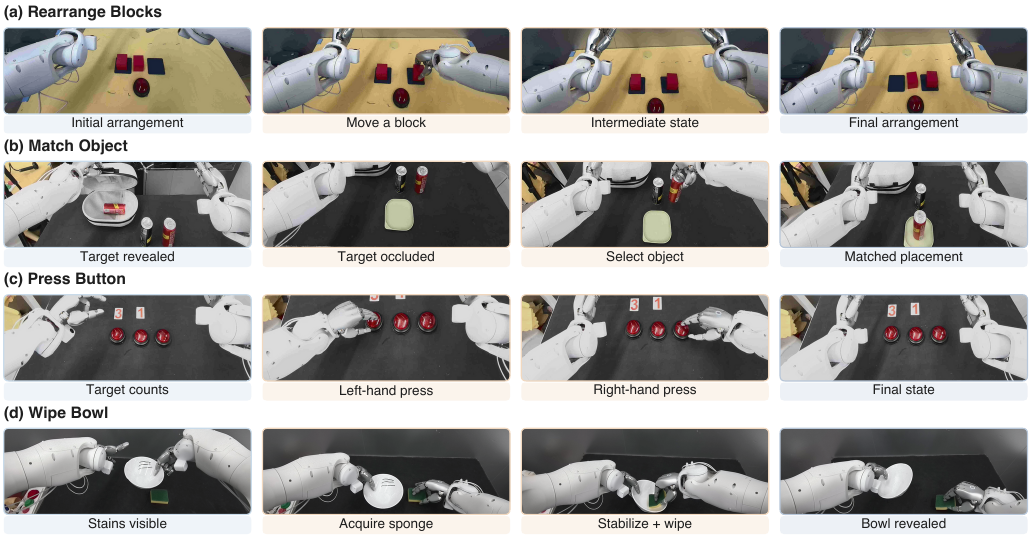}
  \caption{\textbf{Four-stage task workflows on a bimanual platform with dual 7-DoF arms and five-finger dexterous hands.} Chronological head-camera keyframes for (a)~\emph{Rearrange Blocks}, (b)~\emph{Match Object}, (c)~\emph{Press Button}, and (d)~\emph{Wipe Bowl}, corresponding to the step-by-step evaluation procedures in \cref{sec:supp_real_cost} and results in \cref{tab:supp_real}.}
  \label{fig:supp_real}
\end{figure}

\paragraph{Hardware platform, online execution loop, and task workflows.}
Physical evaluations run on a bimanual robot (\cref{fig:supp_real}) combining dual 7-DoF arms, two high-DoF five-finger dexterous hands with 22 joint coordinates per hand, and torso and head actuation across 65 state and action dimensions normalized by 1st and 99th percentiles with two unused base channels masked, observed at 30\,Hz by a cropped $512\times640$ left head-stereo RGB camera and two $384\times480$ wrist fisheye RGB cameras.
Both \method and the matched No-memory baseline initialize from the same task-pretrained 16-layer DiT GR00T~N1.7 policy, with gated memory adapters attached at DiT blocks $\{0,4,8,12\}$ for \method.
At the start of every trial, object placements or target digit cards are randomized under matched distributions across interleaved baseline and \method rollouts, and all Short and Long fast-weight states reset to $W_0$ and $W^{\mathrm L}_0$.
At each policy call, the robot reads memory to predict a 40-step relative joint-angle chunk. It then writes the current $4\times4$ head-camera feature grid and up to 10 normalized joint targets sampled from commands issued between the previous and current observations.

\Cref{tab:supp_real} reports trial counts, success rates, and two-sided 95\% binomial Wilson intervals matching Fig.~\ref{fig:real} of the main paper across the four workflows in \cref{fig:supp_real}a--d:
\begin{itemize}[leftmargin=*,nosep,topsep=2pt]
  \item \textbf{Rearrange Blocks (20 trials/method; \cref{fig:supp_real}a):} One red block starts on a randomly chosen left or right mat and an identical block starts between them behind a central dome button (\emph{Initial arrangement}). The robot (1)~moves the center block onto the empty mat (\emph{Move a block}), creating a visually symmetric two-mat state (\emph{Intermediate state}); (2)~presses the central button; and (3)~relocates the block that \emph{originally} occupied a mat back to the center (\emph{Final arrangement}).
  \item \textbf{Match Object (20 trials/method; \cref{fig:supp_real}b):} The robot places a red or black can from the table into the box. Once the box is closed, the robot selects a can of the same color from the table and places it on the tray.
  \item \textbf{Press Button (45 trials/method; \cref{fig:supp_real}c):} Three red dome buttons sit in front of two digit cards specifying target counts $(c_{\mathrm L}, c_{\mathrm M})$ (\emph{Target counts}). The robot (1)~reads both counts; (2)~presses the left button $c_{\mathrm L}$ times and middle button $c_{\mathrm M}$ times (\emph{Left-hand press}); and (3)~presses the right confirmation button once (\emph{Right-hand press}) before retracting (\emph{Final state}). Success requires the exact sequence of counts $(c_{\mathrm L}, c_{\mathrm M}, 1)$.
  \item \textbf{Wipe Bowl (20 trials/method; \cref{fig:supp_real}d):} The robot steadies a stained bowl with its left hand and wipes the inside with a sponge in its right hand. The strokes repeat up and down. The task tests whether the robot advances to the next area after wiping one spot. A policy that loses track of its progress can keep repeating the same stroke in place. Once the bowl is clean, the robot returns the sponge.
\end{itemize}

\begin{table}[htbp]
  \centering
  \small
  \caption{\textbf{Real-robot manipulation results.} Entries report $k/n$ (\% and two-sided 95\% binomial Wilson CI), with gains in percentage points.}
  \label{tab:supp_real}
  \begin{tabular}{@{}l cc c@{}}
    \toprule
    \headrow Task & No memory & \fullmark\ \method & Gain \\
    \midrule
    Rearrange & 7/20 (35.0; [18.1, 56.7]) & \methodcell{18/20 (90.0; [69.9, 97.2])} & \gainmark{+55.0} \\
    Match Obj. & 6/20 (30.0; [14.5, 51.9]) & \methodcell{17/20 (85.0; [64.0, 94.8])} & \gainmark{+55.0} \\
    Press Btn. & 4/45 (\phantom{0}8.9; [\phantom{0}3.5, 20.7]) & \methodcell{23/45 (51.1; [37.0, 65.0])} & \gainmark{+42.2} \\
    Wipe Bowl & 12/20 (60.0; [38.7, 78.1]) & \methodcell{17/20 (85.0; [64.0, 94.8])} & \gainmark{+25.0} \\
    \midrule
    4-task mean & 33.5\% & \methodcell{77.8\%} & \gainmark{+44.3} \\
    All 105 & 29/105 (27.6; [20.0, 36.8]) & \methodcell{75/105 (71.4; [62.2, 79.2])} & \gainmark{+43.8} \\
    \bottomrule
  \end{tabular}
\end{table}

\paragraph{Training step cost and online memory overhead.}
On RMBench, visual Short pretraining takes 1.59\,h on one H20 GPU (5.65\,GiB peak memory), and visual Long pretraining takes 5.03\,h on eight H20 GPUs at 12 episodes per GPU. Action Short and Long pretraining take 1.61\,h and 0.31\,h, respectively, on four H20 GPUs.
Building each task's offline cache takes 3.9--24.5\,min. Across the nine tasks, Stage~3 policy adaptation at global batch 256 takes a median of 6.80\,h per task on four H20 GPUs, summing logged training segments including work repeated after restarts.
\Cref{tab:supp_cost} and \cref{fig:supp_efficiency}b profile synchronized Stage~3 steps on one H20 GPU across 1.434B active parameters under BF16 autocast with FP32 master and fast weights, reporting medians over 12 steps after 3 warmup steps: at 64 writes and batch 8, cached steps are $7.07\times$ faster ($174.1$ vs.\ $1{,}230.0$\,ms) with 35.5\% lower peak memory ($27.5$ vs.\ $42.6$\,GiB) than joint writer backpropagation over 21.62M writer parameters, and at batch 64 run in 505\,ms per step versus 2,541\,ms for detached prefix replay while joint backpropagation runs out of memory at write 36.

\begin{table}[htbp]
  \centering
  \small
  \caption{\textbf{Stage~3 policy optimization step cost} for GR00T~N1.7 on a single H20 GPU.}
  \label{tab:supp_cost}
  \begin{tabular}{@{}r rrrr@{}}
    \toprule
    & \multicolumn{2}{c}{Step time (ms)} & \multicolumn{2}{c}{Peak GPU (GiB)} \\
    \cmidrule(lr){2-3}\cmidrule(l){4-5}
    \headrow History & Cached & Joint & Cached & Joint \\
    \midrule
    16 writes & 165.1 & 431.8 & 27.5 & 27.8 \\
    64 writes & 174.1 & 1,230.0 & 27.5 & 42.6 \\
    \bottomrule
  \end{tabular}
\end{table}

We measure online storage and latency over 18 RMBench episodes (72 GPU runs), excluding the shared 85.96\,MiB model from dynamic storage. On-demand merging uses 28.51\,MiB versus 46.57\,MiB for immediate merging (38.8\% less). Across 450 demonstrations, both trees cover the full history; the 12-slot sliding buffer uses 10.32\,MiB and covers 69.3\%. For queries 65--256 writes in the past, on-demand merging reduces recall MSE from 0.0873 to 0.0670 (\cref{fig:supp_efficiency}c). Median read and write latencies for the on-demand tree are 0.034 and 7.86\,ms, respectively; the 99th-percentile write latency including warmup is 88.3\,ms.
Retraining 40k-step GR00T~N1.7 policies over 100 trials per task confirms closed-loop retention: trees with on-demand or immediate merging both reach 96--100\% on \emph{Cover Blocks}, \emph{Swap Blocks}, \emph{Battery}, and \emph{Blocks Ranking}, whereas a 12-slot sliding buffer drops to 36\% on \emph{Cover Blocks} and 57\% on \emph{Swap Blocks} once early cues are evicted.

\end{document}